\pdfoutput=1
\documentclass[11pt]{article}

\usepackage[final]{acl}

\usepackage{times}
\usepackage{latexsym}

\usepackage[T1]{fontenc}
\usepackage[utf8]{inputenc}

\usepackage{microtype}

\usepackage{inconsolata}

 \usepackage{makecell, array}
 \usepackage{multirow}
\usepackage{graphicx}
\usepackage{algorithm}
\usepackage{amsfonts}
\usepackage{algorithmic}
\usepackage{amsmath}
\usepackage{booktabs}
\usepackage{caption} % Add to preamble
\usepackage{subcaption}
\usepackage[table]{xcolor} % in preamble
\usepackage{cleveref} %load it at the end

\usepackage{url}
\newcommand\blfootnote[1]{%
  \begingroup
  \renewcommand\thefootnote{}%
  \footnote{#1}%
  \addtocounter{footnote}{-1}%
  \endgroup
}

\newcommand{\mycc}{\cellcolor{gray!40}}
\title{LLM-Guided Co-Training for Text Classification}

\author{Md Mezbaur Rahman \\
  Computer Science \\
  University of Illinois Chicago \\
  \texttt{\color{blue}mrahma56@uic.edu} \\\And
  Cornelia Caragea \\
  Computer Science \\
  University of Illinois Chicago \\
  \texttt{\color{blue}cornelia@uic.edu} \\}

\begin{document}
\maketitle
\begin{abstract}

In this paper, we introduce a novel weighted co-training approach that is guided by Large Language Models (LLMs).
Namely, in our co-training approach, we use LLM labels on unlabeled data as target labels and co-train two encoder-only based networks that train each other over multiple iterations: first, all samples are forwarded through each network and historical estimates of each network's confidence in the LLM label are recorded; second, a dynamic importance weight is derived for each sample according to each network's belief in the quality of the LLM label for that sample; finally, the two networks exchange importance weights with each other---each network back-propagates all samples weighted with the importance weights coming from its peer network and updates its own parameters. By strategically utilizing LLM-generated guidance, our approach significantly outperforms conventional SSL methods, particularly in settings with abundant unlabeled data.  Empirical results show that it achieves state-of-the-art performance on 4 out of 5 benchmark datasets and ranks first among 14 compared methods according to the Friedman test. Our results highlight a new direction in semi-supervised learning---where LLMs serve as knowledge amplifiers, enabling backbone co-training models to achieve state-of-the-art performance efficiently. 
\blfootnote{We make our implementation publicly available at \url{https://github.com/mezbaur-rahman/Lg-CoTrain}}
\end{abstract}

\section{Introduction}

Semi-Supervised Learning (SSL) has gained substantial attention for its ability to improve model performance by leveraging a small set of labeled data alongside a large pool of unlabeled data \cite{wang2022usb, van2020survey}. A widely used SSL strategy is pseudo-labeling that relies on the model itself to generate pseudo-labels for unlabeled data and uses these pseudo-labels during training. FixMatch exemplifies this by selecting high-confidence predictions exceeding a fixed threshold as pseudo-labels. However, this rigid threshold often excludes diverse examples with lower confidence scores. To address this, recent methods either: (1) adaptively adjust class-wise thresholds to incorporate more examples \cite{zhang2021flexmatch}, or (2) assign confidence-based weights to all unlabeled samples \cite{chen2023softmatch}. While these strategies improve data utilization, they remain dependent on the model's own potentially unreliable predictions, introducing noisy pseudo-labels that can harm generalization, especially due to overconfident yet incorrect predictions.

Recent advances have highlighted the potential of LLMs in producing high-quality pseudo-labels for SSL. The VerifyMatch framework \cite{park-caragea-2024-verifymatch} incorporates LLM-generated pseudo-labels into the SSL setup and tackles their noise by first identifying noisy samples using a verifier network. The verifier flags low-confidence pseudo-labels (based on softmax probability) and applies MixUp \cite{zhang2018mixup} to blend them with human-annotated data, thereby mitigating noise and improving label quality. Although this approach presents a strong case in combining LLM-generated pseudo-labels into the SSL setup, it uses only the confidence scores from the current training iteration from a single verifier model to distinguish between potentially correct and noisy samples. This limited view misses opportunities to capture richer training dynamics across time or across multiple models equipped with complementary knowledge and learning abilities, which can further enhance the robustness of pseudo-labeling.

In this paper, we propose a novel co-training framework that leverages LLM-generated pseudo-labels without modifying or discarding any samples. Our approach diverges from VerifyMatch \cite{park-caragea-2024-verifymatch}, which mitigates noise by altering low-confidence pseudo-labels (e.g., via MixUp with human-labeled data). Instead, we preserve all pseudo-labels and reweight them based on their estimated reliability, allowing the model to learn from the entire data while reducing the influence of incorrect pseudo-labels. Moreover, while VerifyMatch uses a single verifier model to estimate the label quality, we employ a dual model-based weighting mechanism. Our framework also diverges from traditional co-training methods \cite{blum1998combining} that exchange the most confident pseudo-labels between models; we instead exchange importance weights that reflect each model’s belief in the quality of the LLM pseudo-labels. Leveraging the insights from \citet{swayamdipta-etal-2020-dataset}, our framework goes beyond relying solely on confidence from the current training epoch. Instead, it utilizes historical training dynamics and the consistency and disagreements of the two models' predictions across epochs to assign higher weights to likely correct samples and lower weights to potentially incorrect ones.

Our contributions are as follows:
\begin{itemize}
    \item We propose \textsc{LG-CoTrain}, a novel co-training method that leverages pseudo-labels generated by LLMs alongside a small set of human-annotated labels. Two peer networks are trained jointly, each guiding the other using its training dynamics to maintain divergence and promote complementary learning.
\vspace{-6mm}
    \item \textsc{LG-CoTrain} consistently outperforms all standard SSL baselines across five benchmark classification datasets and exceeds the performance of the LLMs themselves on four out of five datasets. Remarkably, it also achieves the top-1 rank in the Friedman test among 14 methods, reducing the mean error rate by 1.56\% compared to the second-best baseline.
\vspace{-2mm}
    \item We conduct extensive ablation studies and qualitative analyses to validate the effectiveness of our proposed method in both balanced and imbalanced data distributions. %, %demonstrating their individual contributions to the overall performance. 
    We also evaluate the effectiveness of \textsc{LG-CoTrain} in several natural language understanding tasks.

\end{itemize}

\section{Related Works}

\subsection{Large Language Models}

Large Language Models (LLMs) have demonstrated remarkable capabilities across a wide range of NLP tasks, including text generation, question answering, summarization, and dialogue generation \cite{zhang-etal-2020-dialogpt, touvron2023llama, chen-etal-2024-flexiqa}. Their generalization ability and scalability have made them a key component in modern NLP pipelines. Earlier approaches such as BERT \cite{devlin-etal-2019-bert} relied on task-specific fine-tuning, requiring labeled data and weight updates for each new task. However, as LLMs have grown in size and training scope, prompting has emerged as a more flexible alternative \cite{NEURIPS2020_1457c0d6, wei2022finetuned}. Prompting enables LLMs to perform tasks by conditioning on task descriptions or a few in-context examples written in natural language, allowing for zero-shot and few-shot learning. % without modifying model parameters. 
In our work, we leverage LLM prompting to generate high-quality pseudo-labels for unlabeled data. We also include prompting-based baselines in our evaluations to assess the effectiveness of our method relative to direct LLM predictions.

\subsection{Semi Supervised Learning}

Semi-Supervised Learning (SSL) has led to a surge of research based on self-training and pseudo-labeling \cite{xie2020self, sohn2020fixmatch, sosea-caragea-2022-leveraging, sadat-caragea-2022-learning, sosea2023marginmatch, chen2023softmatch, zou-caragea-2023-jointmatch, zou-etal-2023-decrisismb,gyawali-etal-2024-gunstance, hosseini-caragea-2023-semi}, where models are trained using a mix of labeled and pseudo-labeled data. Popular methods like FixMatch \cite{sohn2020fixmatch} and FlexMatch \cite{zhang2021flexmatch} rely on confidence thresholding to select pseudo-labels, but it often results in discarding a large portion of the unlabeled data. To improve data utilization, \citet{chen2023softmatch} proposed SoftMatch, which retains all samples by assigning lower weights to low-confidence samples during training. However, whether through hard thresholding or confidence-based weighting, methods that rely solely on a single model for pseudo-labeling remain vulnerable to assigning high weights to incorrect pseudo-labels, especially in the early stages of training.

\subsection{Learning with Noisy Labels}

The challenge of learning from noisy labels has been explored for decades \cite{angluin1988learning}, with renewed focus in deep learning due to neural networks' tendency to memorize mislabeled data \cite{arpit2017closer}. Noisy labels—corrupted deviations from ground truth labels—can significantly degrade model performance. To combat this, several strategies have been proposed to distinguish between reliable and unreliable training samples. MentorNet \cite{jiang2018mentornet} proposed an auxiliary network to select clean samples, but its reliance on a single model introduces confirmation bias. Co-teaching \cite{han2018co} addressed this by training two peer networks that exchange low-loss samples, minimizing mutual errors. DivideMix \cite{lidividemix} treated noisy label learning as a semi-supervised task, using Gaussian Mixture Models to split data into confident and unconfident subsets and co-trained two models on complementary partitions. VerifyMatch \cite{park-caragea-2024-verifymatch} used LLM-generated pseudo-labels, which are generally of high quality but still noisy. VerifyMatch introduces a verifier network to estimate label confidence: low-confidence, noisy samples are refined using MixUp \cite{zhang2018mixup} and human-annotated data, while high-confidence samples are used directly. %in supervised training.

In contrast, we propose a dual-model semi-supervised framework that integrates LLM-generated pseudo-labels while training two classifiers in parallel. Unlike VerifyMatch \cite{park-caragea-2024-verifymatch}, which employs a single-model setup with a verifier, our approach leverages a dual-model architecture and dynamically weights training samples based on cross-model training dynamics. Another key distinction is our use of multi-epoch training dynamics to assess model behavior consistency over time—an aspect VerifyMatch overlooks. Furthermore, our method differs from DivideMix \cite{lidividemix} and Co-Teaching \cite{han2018co}, which rely on label exchange or sample selection during training. Instead, we retain all pseudo-labels without discarding or modifying them, and guide the training process through historical training dynamics based weighting. This strategy encourages model divergence and enhances robustness while preserving the integrity of the pseudo-labeled data.

\section{Proposed Approach}
\label{sec_4}

% In this section, we present \textsc{LG-CoTrain}, our proposed LLM-Guided Weighted Co-Training framework designed in a semi-supervised setup.

In this section, we present \textsc{LG-CoTrain}, our proposed approach for semi-supervised learning. We begin by defining the notation used throughout the method. We then describe the pseudo-label generation process using LLMs (\S\ref{sec:4_1}). Next, we describe the weighting metrics to estimate the reliability of each pseudo-labeled sample (\S\ref{sec:4_2}). Finally, we present our weighted co-training strategy, where two peer models are trained jointly using these weights, which are continuously updated during training (\S\ref{sec:4_3}).

\textbf{Notation.}  
We consider a training dataset \( D \) consisting of a small labeled set \(\mathcal{D}_l = \{(x_i^l, y_i^l)\}_{i=1}^{N_l}\) and a much larger unlabeled set \(\mathcal{D}_u = \{x_j^u\}_{j=1}^{N_u}\). Here, \(x\) denotes an input sample (e.g., a text sequence), and \(y \in \{1, \dots, C\}\) denotes its corresponding class label. We define \(N_l\), \(N_u\), and \(C\) as the number of labeled samples, unlabeled samples, and class categories, respectively, where \(N_u \gg N_l\).

\subsection{Pseudo-Label Generation with LLMs}  
\label{sec:4_1}

We utilize LLMs to generate pseudo-labels for unlabeled texts through task-specific prompting designed for classification. We use the LLMs in zero-shot or few-shot settings to generate pseudo-labels for the unlabeled samples.

% The LLM is queried in a zero-shot or few-shot setting using these prompts. 

% For each class \( c \), we define a corresponding label word and construct classification prompts tailored to the target task. 

Given an unlabeled input \( x_j^u \in \mathcal{D}_u \), the prompt \( P(x_j^u) \) formats the input into a query suitable for the LLM. The model then generates an output token sequence, from which a label word is extracted and mapped to class \( c \), resulting in the pseudo-label \( \tilde{y}_j = c \). This yields the pseudo-labeled set:
\[
\mathcal{D}_{LG} = \{(x_j^u, \tilde{y}_j) \mid x_j^u \in \mathcal{D}_u, \tilde{y}_j \in \{1, \dots, C\}\}
\]

This process ensures that \(\mathcal{D}_{LG}\) contains samples with valid class-aligned pseudo-labels. These samples serve as additional supervision during the co-training stage. The prompt templates and the prompting strategies are provided in Appendix~\ref{apx:prompts}.

\subsection{Weight Generation}
\label{sec:4_2}

The two networks (i.e., the classifiers) in our co-training setup are denoted as \(\theta_1\) and \(\theta_2\). At each training epoch, we assign importance weights to the samples from \(\mathcal{D}_{LG}\) based on the training dynamics of each classifier. The goal is twofold: (1) to emphasize reliable pseudo-labeled examples, and (2) to maintain divergence between the classifiers, enabling them to learn complementary signals and suppress the effects of noisy supervision.

To assess sample reliability, we make use of two metrics—\textbf{confidence} and \textbf{variability}—as proposed in Dataset Cartography~\cite{swayamdipta-etal-2020-dataset}. These metrics capture the historical behavior of a model on each sample during training. \textbf{Confidence} reflects how confidently a model predicts the assigned label over iterations. For example, for a sample \((x, \tilde{y})\), the confidence of a classifier \(\theta\) across \(e\) training iterations is computed as:
\vspace{-2mm}
\begin{equation}
c_\theta(x, \tilde{y}) = \frac{1}{e} \sum_{t=1}^{e} p(\tilde{y} \vert x; \theta)
\label{eq1}
\end{equation}

\noindent
\textbf{Variability} captures how consistently a model predicts the assigned label over iterations, i.e., the standard deviation of these predictions, indicating how stable the model’s belief in the label is:
\vspace{-2mm}
\begin{equation}
v_\theta(x, \tilde{y}) = \sqrt{\frac{1}{e} \sum_{t=1}^{e} \left(p(\tilde{y} \vert x; \theta_t) - c_\theta(x, \tilde{y})\right)^2}
\label{eq2}
\end{equation}
% \begin{equation}
% v_\theta(x, \tilde{y}^{\text{llm}}) = \sqrt{\textstyle \frac{1}{e} \sum_{t=1}^{e} \left(p(\tilde{y}^{\text{llm}} \vert x; \theta_t) - c_\theta(x, \tilde{y}^{\text{llm}})\right)^2}
% \label{eq2}
% \end{equation}

While these metrics individually signal sample quality, they do not directly promote complementary learning between the two classifiers. To address this, inspired by prior works \cite{poesina-etal-2024-novel, sadat-caragea-2024-co}, we use two asymmetric importance weighting schemes—\(\lambda_1\) and \(\lambda_2\)—which leverage the confidence and variability of each sample in opposing ways. These weights are used to supervise the \textit{peer} network, thereby encouraging divergence and suppressing mutual error reinforcement. Specifically, for each pseudo-labeled sample \((x^i, \tilde{y}^i)\) and each classifier, \(\theta_1\) and \(\theta_2\), we define:

\vspace{-2mm}
\begin{equation}
\lambda_1^i = c_{\theta_1}(x^i, \tilde{y}^i) + v_{\theta_1}(x^i, \tilde{y}^i)
\label{eq3}
\end{equation}

\vspace{-2mm}
\begin{equation}
\lambda_2^i = c_{\theta_2}(x^i, \tilde{y}^i) - v_{\theta_2}(x^i, \tilde{y}^i)
\label{eq4}
\end{equation}

% This formulation causes \(\theta_1\) to emphasize samples with either strong or uncertain signals, while \(\theta_2\) focuses on samples with both high confidence and stability. The result is a natural divergence in sample preferences across models.

As a result, examples that are easy to classify by both classifiers (i.e., high confidence and low variability) produce high values for both \(\lambda_{1}^{i}\) and \(\lambda_{2}^{i}\). Conversely, hard examples (i.e., low confidence and low variability) yield low values for both \(\lambda_{1}^{i}\) and \(\lambda_{2}^{i}\). For ambiguous examples, defined by high variability and moderate confidence by both classifiers, the weights differ between the classifiers. Specifically, the weight for \(\theta_1\) (i.e., \(\lambda_{2}^{i}\)) is low due to the classifiers showing only moderate confidence in these examples. That is, a confidence of 0.5 and a variability of 0.4 result in a low weight of 0.1. In contrast, the weight for \(\theta_2\) (i.e., \(\lambda_{1}^{i}\)) remains high even with moderate confidence, when the variability is high. That is, a confidence of 0.5 and variability of 0.4 result in a high weight of 0.9.
 
% To ensure consistency, we normalize the weights using a min-max normalization method, scaling both sets of weights to the range \([0, 1]\). 
% This weighting strategy ensures that easy examples, which are more likely to be correctly labeled, are heavily weighted in the training of both classifiers. Conversely, hard examples, which are more prone to incorrect labeling, are assigned low weights, minimizing their influence on both classifiers. 

This weighting strategy gives higher weight to easy, likely correct examples, while down-weighting harder, noisier ones to reduce their impact on both classifiers. For ambiguous examples, where the classifiers exhibit uncertainty about the assigned labels, one classifier receives high weights for these examples, while the other receives low weights. This approach preserves divergence between the classifiers and allows them to learn complementary information from the data.

\begin{algorithm}[H] % Use [H] to force positioning
\small 
\caption{\textsc{LG-CoTrain}}
\label{alg:weighted_cotraining}
        \begin{algorithmic}[1]
           % \STATE {\bfseries Input:} Labeled set $\mathcal{D}_l = \{(\mathbf{x}^i, y^i)\}_{i=1}^{N_l}$,\\ 
           % Unlabeled set $\mathcal{D}_{u} = \{\mathbf{x}^i\}_{i=1}^{N_u}$\\
           % Two classifier models  $\theta_1$, $\theta_2$\\

           % Probability matices $\mathcal{P}_1, \mathcal{P}_2 \in \mathbb{R}^{T \times |D_{LG}|}$\\
           %  $\quad \mathcal{P}_1 = \{p_t(y^i | \mathbf{x}^i; \theta_1)\}_{t,i}$, 
           %  $\: \mathcal{P}_2 = \{p_t(\tilde{y}^i | \mathbf{x}^i; \theta_2)\}_{t,i}$

          \STATE {\bfseries Input:} Labeled set $\mathcal{D}_l = \{(\mathbf{x}^i, y^i)\}_{i=1}^{N_l}$; \\
         Unlabeled set $\mathcal{D}_u = \{\mathbf{x}^i\}_{i=1}^{N_u}$; 
         Classifiers $\theta_1$, $\theta_2$; \\
         Weight generation epochs $T$, Max co-training epochs $E$; \\
         Probability matrices $\mathcal{P}_1, \mathcal{P}_2 \in \mathbb{R}^{T \times |\mathcal{D}_{LG}|}$

           \STATE Construct $\mathcal{D}_{LG} = \{(\mathbf{x}^i, \tilde{y}^i)\}_{i=1}^{N_{LG}}$ $\leftarrow$ Sec.~\ref{sec:4_1}
           \STATE Divide $\mathcal{D}_l$ into two balanced subsets $\mathcal{D}_{l_1}$ and $\mathcal{D}_{l_2}$

           \FOR{iter \( t \in \{1, 2, \dots, T\} \)}
                \STATE Train \( \theta_1 \) and \( \theta_2 \) using \( \mathcal{D}_{l_1} \) and \( \mathcal{D}_{l_2} \), respectively;

                % \STATE for all \( i \in \mathcal{D}_{LG} \) Store\\
                % \( \mathcal{P}_1[t, i] = p_t(y^i | \mathbf{x}^i; \theta_1) \) and \( \mathcal{P}_2[t, i] = p_t(\tilde{y}^i | \mathbf{x}^i; \theta_2) \)

                \FOR{each \( i \in \mathcal{D}_{LG} \)}
                    \STATE $\mathcal{P}_1[t, i] \leftarrow p_t(y^i \mid \mathbf{x}^i; \theta_1)$
                    \STATE $\mathcal{P}_2[t, i] \leftarrow p_t(\tilde{y}^i \mid \mathbf{x}^i; \theta_2)$
                \ENDFOR
            \ENDFOR

           \FOR{each sample $i \in D_{LG}$}
                \STATE $c_{\theta_1}^i$, $c_{\theta_2}^i$ $\leftarrow$  \textsc{GetConf}$(\mathcal{P}_1, i)$,  
                \: \textsc{GetConf}$(\mathcal{P}_2, i)$ \\
                $v_{\theta_1}^i$, $v_{\theta_2}^i$ $\leftarrow$  \textsc{GetVar}$(\mathcal{P}_1, i)$, \:  \textsc{GetVar}$(\mathcal{P}_2, i)$
                
                \STATE $\lambda_1^i \leftarrow c_{\theta_1}^i + v_{\theta_1}^i$;\quad
                $\lambda_2^i \leftarrow c_{\theta_2}^i - v_{\theta_2}^i$. 
                
                % \hfill {\color{gray}\textbf{\texttt{\#}} initial weights}

            \ENDFOR

        \STATE Re-initialize $\theta_1$, $\theta_2$. 
        
        % \hfill {\color{gray} \textbf{\texttt{\#}} Re-init classifiers for co-training.}
        \FOR{iter \( \epsilon \in \{1, 2, \dots, E\} \)}
            \FOR{each mini-batch $B \in D_{LG}$}
                % \STATE {\color{gray} \textbf{\texttt{\#}} Calculate cross-entropy loss.}\\
                \STATE $\mathcal{L}_1 \leftarrow \frac{1}{|B|} \sum_{i=0}^{|B|} \lambda_2^i * H(\tilde{y}^i, p_d(\mathbf{x}^i;\theta_1))$ 
                \STATE $\mathcal{L}_2 \leftarrow \frac{1}{|B|} \sum_{i=0}^{|B|} \lambda_1^i * H(\tilde{y}^i, p_d(\mathbf{x}^i;\theta_2))$ 
                \FOR{each example $i \in B$}
                    % \STATE Update $c_{\theta_1}(\mathbf{x}^i, \tilde{y}^i)$ and $c_{\theta_2}(\mathbf{x}^i, \tilde{y}^i)$ $\leftarrow$ Eq. ~\eqref{eq1}; 
                    % \STATE Update $v_{\theta_1}(\mathbf{x}^i, \tilde{y}^i)$ and $v_{\theta_2}(\mathbf{x}^i, \tilde{y}^i)$ $\leftarrow$ Eq. ~\eqref{eq2}.
                    \STATE Update $c_{\theta_k}^i$, $v_{\theta_k}^i$ via Eq.~\eqref{eq1},~\eqref{eq2} for $k=1,2$\;
                    
                    \STATE $\lambda_1^i \leftarrow c_{\theta_1}(\mathbf{x}^i, \tilde{y}^i) + v_{\theta_1}(\mathbf{x}^i, \tilde{y}^i)$\\ 
                    \STATE $\lambda_2^i \leftarrow c_{\theta_2}(\mathbf{x}^i, \tilde{y}^i) - v_{\theta_2}(\mathbf{x}^i, \tilde{y}^i)$ 
                    
                    % \hfill {\color{gray} \textbf{\texttt{\#}} Update Importance weights}
                \ENDFOR
            \ENDFOR
            \STATE $\theta_1 \leftarrow \theta_1 - \eta * \nabla \mathcal{L}_1$;\quad $\theta_2 \leftarrow \theta_2 - \eta * \nabla \mathcal{L}_2$. 
            
            % \hfill {\color{gray} \textbf{\texttt{\#}} Update the parameters of $\theta_1$, and $\theta_2$}
        \ENDFOR

        \STATE Fine-tune $\theta_1$ and $\theta_2$ using $\mathcal{D}_{l_1}$ and $\mathcal{D}_{l_2}$ respectively

        \end{algorithmic}
\end{algorithm}

\subsection{Weighted Co-Training}
The weighted co-training module consists of three main steps, which are outlined below:

% \textbf{Step 1: Initial Weight Assignment}  
\paragraph{Initial Weight Assignment}
To initialize the importance weights for the samples in the pseudo-labeled set \( \mathcal{D}_{LG} \), the classifiers \( \theta_1 \) and \( \theta_2 \) are first trained on different, equally sized subsets of the labeled dataset \( \mathcal{D}_{l} \), denoted as \( \mathcal{D}_{l_1} \) and \( \mathcal{D}_{l_2} \), respectively. These subsets are constructed to maintain similar class distributions, ensuring a fair training process.

Next, we define two probability matrices, \( \mathcal{P}_1, \mathcal{P}_2 \in \mathbb{R}^{T \times |\mathcal{D}_{LG}|} \), where each entry stores the predicted probabilities \( p(\tilde{y}_i \mid x_i; \theta_1) \) and \( p(\tilde{y}_i \mid x_i; \theta_2) \) for each sample \( i \) in \( \mathcal{D}_{LG} \) across epochs \( t = 1, \dots, T \). These probabilities are recorded at each training epoch.

For each sample in \( \mathcal{D}_{LG} \), we compute confidence and variability using Eq.~\eqref{eq1} and Eq.~\eqref{eq2}, respectively, based on the corresponding probability matrix. These values are then substituted into Eq.~\eqref{eq3} and Eq.~\eqref{eq4} to determine the initial importance weights $\lambda_{1}$ and $\lambda_{2}$.

% \textbf{Step 2: Co-Training}

% In this step, we begin by re-initializing the two classifiers, \(\theta_1\) and \(\theta_2\). Both classifiers are then co-trained using the pseudo-labeled dataset \( \mathcal{D}_{LG} \). After each training epoch, the cross-entropy loss for each classifier is scaled by the importance weights assigned by the other classifier. 
\paragraph{Co-Training}
\label{sec:4_3}
In this step, we begin by re-initializing the two classifiers, \(\theta_1\) and \(\theta_2\). Both classifiers are then co-trained using the pseudo-labeled dataset \(\mathcal{D}_{LG}\). After each training epoch, for model \(\theta_1\), each sample’s loss is scaled by the weight metric \(\lambda_2\); similarly, for model \(\theta_2\), each sample’s loss is scaled by the weight metric \(\lambda_1\). Specifically, for each mini-batch \(B \subseteq \mathcal{D}_{LG}\), the losses \(\mathcal{L}_1\) and \(\mathcal{L}_2\) for \(\theta_1\) and \(\theta_2\), respectively, are computed as follows:

\vspace{-5mm}
\begin{equation}
\mathcal{L}_1 = \frac{1}{|B|} \sum_{i=1}^{|B|} \lambda^i_{2} \ast H(\tilde{y}^i, p_d(x^i; \theta_1)).
\end{equation}
\vspace{-3mm}
\begin{equation}
\mathcal{L}_2 = \frac{1}{|B|} \sum_{i=1}^{|B|} \lambda^i_{1} \ast H(\tilde{y}^i, p_d(x^i; \theta_2)).
\end{equation}

Here, \(p_d\) represents the probability distribution over the labels predicted by a classifier for a pseudo-labeled sample \(x^i\) in \(B\), and \(H\) denotes the standard cross-entropy loss. 

After calculating the loss for both classifiers, the confidence and variability metrics are updated using Eq.~\eqref{eq1} and Eq.~\eqref{eq2}. It is important to note that the probabilities predicted by the classifiers during the final epoch of Step 1 are used as the initial probabilities for both classifiers. These initial probabilities are included in the calculation of the mean and standard deviation throughout the co-training epochs. Subsequently, both sets of weights are recalculated using Eq.~\eqref{eq3} and Eq.~\eqref{eq4}, based on the updated confidence and variability metrics, to be applied in the next epoch. 

This iterative process allows the classifiers to co-train by exchanging complementary information derived from their respective training dynamics across epochs. Crucially,  each model guides the other by dynamically adjusting the influence of each training sample using the weight metric $\lambda$.

% \textbf{Step 3: Fine-tuning} 
\paragraph{Fine-tuning}
Upon completing the co-training process after a predefined maximum of \(E\) epochs, the classifiers \(\theta_1\) and \(\theta_2\) are fine-tuned using the labeled subsets \( \mathcal{D}_{l_1} \) and \( \mathcal{D}_{l_2} \), respectively.

During inference, we ensemble the two models by averaging their softmax outputs and taking the argmax to get the final prediction.

\section{Experiments}
In this section, we describe the datasets used for evaluation (\S\ref{datasets}), present our experimental setup and baselines (\S\ref{expsetup}), and discuss the results (\S\ref{results}).

\subsection{Datasets}
\label{datasets}
We conduct experiments on five widely used classification datasets: IMDB \cite{maas-etal-2011-learning}, AG News \cite{NIPS2015_250cf8b5}, Yahoo! Answers \cite{chang2008importance}, Amazon Reviews \cite{mcauley2013hidden}, and Yelp Reviews \cite{yelp_dataset}, along with three natural language understanding (NLU) tasks: QQP \cite{qqp}, SWAG \cite{zellers-etal-2018-swag}, and MNLI \cite{williams-etal-2018-broad}. The first five datasets are used to evaluate performance against SSL baselines, while the remaining three assess generalization on NLU tasks. For the classification datasets, we use the pre-processed splits (labeled, unlabeled, validation, and test) provided by \citet{wang2022usb},\footnote{\url{https://github.com/microsoft/Semi-supervised-learning/tree/main}} released under the MIT license. For the NLU tasks, we use the official dev set for validation and construct the labeled and unlabeled sets by randomly sampling from their respective training sets.
Further dataset details are provided in Appendix~\ref{apx:datasets}.

\subsection{Experimental Setup \& Baselines}
\label{expsetup}
% Our approach leverages pseudo-labels generated by large language models (LLMs) using both zero-shot and few-shot prompting. 

We evaluate our method using both BERT-base and RoBERTa-base as backbone classifiers to demonstrate generality across architectures.  All experiments are conducted with three independent runs, and results are reported using \textit{mean error rate} and standard deviation. In addition, we compute the \textit{Friedman rank} \cite{friedman1940comparison} of each algorithm to assess overall ranking across tasks: \( \text{rank}_F = \frac{1}{m} \sum_{i=1}^{m} \text{rank}_i \), where \( m = 15 \) is the number of evaluation setups and \( \text{rank}_i \) is the algorithm's rank in the \( i \)-th setup. Our implementation details are presented in Appendix~\ref{apx:implementational_details}.

\setlength{\tabcolsep}{1.4pt} % global spacing between columns
\begin{table*}[t]
\centering
\renewcommand{\arraystretch}{1.1} %global vertical spacing between rows
\scriptsize

% \caption{Error rate (\%) and Rank across datasets and label sizes. The first 8 rows are from \citet{wang2022usb}. Subsequent rows show results using pseudo-labels generated by Phi-3 with zero-shot prompting, including our \textbf{LG-CoTr}. Subscripts \textsubscript{R} and \textsubscript{B} indicate RoBERTa and BERT backbones, respectively. Best results are \textbf{bolded}. Superscript \textsuperscript{\textdagger} indicates a statistically significant improvement ($p < 0.05$) over the best baseline using a paired t-test.}

% \vspace{1em}
% \label{tab:results}
\vspace{1em}
\resizebox{0.96\textwidth}{!}{
\begin{tabular}{l|cc|cc|cc|cc|cc|c|c|c}
\toprule
Dataset           & \multicolumn{2}{c|}{IMDB}         & \multicolumn{2}{c|}{AG News}      & \multicolumn{2}{c|}{Amazon Review} & \multicolumn{2}{c|}{Yahoo! Answer} & \multicolumn{2}{c|}{Yelp Review}  & Mean & Fried. & Final \\ 
% \midrule
% \cmidrule(lr){1-1} \cmidrule(lr){2-3} \cmidrule(lr){4-5} \cmidrule(lr){6-7} \cmidrule(lr){8-9} \cmidrule(lr){10-11} \cmidrule(lr){12-14}
\# Label          & 20             & 100             & 40           & 200           & 250             & 1000           & 500             & 2000           & 250             & 1000           & Err.            & rank             & rank           \\ \midrule

Sup (full)  & 
\multicolumn{2}{|c|}{5.69\textsubscript{\tiny{0.15}} } & \multicolumn{2}{|c|}{5.73\textsubscript{\tiny{0.11}} } & \multicolumn{2}{|c|}{36.38\textsubscript{\tiny{0.01}} } & \multicolumn{2}{|c|}{24.86\textsubscript{\tiny{0.07}} } & \multicolumn{2}{|c|}{31.98\textsubscript{\tiny{0.20}} } & 21.07 & - & - 
\\

% supervised       & 20.31\textsubscript{\tiny{2.79}}  & 14.02\textsubscript{\tiny{1.22}}  & 15.06\textsubscript{\tiny{1.08}} & 14.25\textsubscript{\tiny{0.97}} & 52.31\textsubscript{\tiny{1.28}} & 47.53\textsubscript{\tiny{0.69}} & 37.43\textsubscript{\tiny{0.29}} & 33.26\textsubscript{\tiny{0.1}} & 51.22\textsubscript{\tiny{0.98}} & 46.71\textsubscript{\tiny{0.37}} & 31.71 & - & - \\ 
\midrule

% II-Model          & 49.99\textsubscript{\tiny{0.01}}  & 44.75\textsubscript{\tiny{3.99}}  & 60.7\textsubscript{\tiny{19.09}} & 12.58\textsubscript{\tiny{0.57}} & 77.22\textsubscript{\tiny{1.5}} & 53.17\textsubscript{\tiny{2.56}} & 44.91\textsubscript{\tiny{1.32}} & 32.45\textsubscript{\tiny{0.45}} & 75.73\textsubscript{\tiny{4.01}} & 59.82\textsubscript{\tiny{0.61}} & 51.13 & 16 & 16 \\
% \tiny{Ps. Labeling}      & 45.45\textsubscript{\tiny{4.43}}  & 19.67\textsubscript{\tiny{1.01}}  & 19.49\textsubscript{\tiny{3.07}} & 14.69\textsubscript{\tiny{1.88}} & 53.45\textsubscript{\tiny{1.9}} & 47.0\textsubscript{\tiny{0.79}} & 37.7\textsubscript{\tiny{0.65}} & 32.72\textsubscript{\tiny{0.31}} & 54.51\textsubscript{\tiny{0.82}} & 47.33\textsubscript{\tiny{0.2}} & 35.76 & 13 & 13 \\

% \tiny{Mean Teacher}      & 20.06\textsubscript{\tiny{2.51}}  & 13.97\textsubscript{\tiny{1.49}}  & 15.17\textsubscript{\tiny{1.21}} & 13.93\textsubscript{\tiny{0.65}} & 52.14\textsubscript{\tiny{0.52}} & 47.66\textsubscript{\tiny{0.84}} & 37.09\textsubscript{\tiny{0.18}} & 33.43\textsubscript{\tiny{0.28}} & 50.6\textsubscript{\tiny{0.62}} & 47.21\textsubscript{\tiny{0.31}} & 31.63 & 11 & 11 \\

% VAT              & 25.93\textsubscript{\tiny{2.58}}  & 11.61\textsubscript{\tiny{1.79}}  & 14.7\textsubscript{\tiny{1.19}} & 11.71\textsubscript{\tiny{0.84}} & 49.83\textsubscript{\tiny{0.46}} & 46.54\textsubscript{\tiny{0.31}} & 34.87\textsubscript{\tiny{0.41}} & 31.5\textsubscript{\tiny{0.35}} & 52.97\textsubscript{\tiny{1.41}} & 45.3\textsubscript{\tiny{0.32}} & 30.50 & 9 & 9 \\

Sup (lb)               & 
20.31\textsubscript{\tiny{2.79}} & 14.02\textsubscript{\tiny{1.22}} & 15.06\textsubscript{\tiny{1.08}} & 14.25\textsubscript{\tiny{0.97}} & 52.31\textsubscript{\tiny{1.28}} & 47.53\textsubscript{\tiny{0.69}} & 37.43\textsubscript{\tiny{0.29}} & 33.26\textsubscript{\tiny{0.10}} & 
51.22\textsubscript{\tiny{0.98}} & 46.71\textsubscript{\tiny{0.37}} & 
33.31 & - & - \\

Sup (lb+ps) &
10.68\textsubscript{\tiny{1.17}} &
7.88\textsubscript{\tiny{0.01}} &
11.88\textsubscript{\tiny{0.07}} &
10.92\textsubscript{\tiny{0.25}} &
47.77\textsubscript{\tiny{0.53}} &
42.23\textsubscript{\tiny{0.42}} &
33.38\textsubscript{\tiny{0.34}} &
32.15\textsubscript{\tiny{0.52}} &
44.44\textsubscript{\tiny{0.49}} &
39.57\textsubscript{\tiny{0.23}} &
28.09 & 9.8 & 11\\
\midrule

UDA              & 49.97\textsubscript{\tiny{0.04}}  & 50.0\textsubscript{\tiny{0.0}}  & 41.0\textsubscript{\tiny{24.96}} & 53.68\textsubscript{\tiny{30.15}} & 60.76\textsubscript{\tiny{13.61}} & 68.38\textsubscript{\tiny{16.44}} & 71.3\textsubscript{\tiny{26.45}} & 70.5\textsubscript{\tiny{27.58}} & 69.33\textsubscript{\tiny{15.08}} & 66.95\textsubscript{\tiny{18.46}} & 60.19 & 15 & 15 \\
 
Dash             & 8.34\textsubscript{\tiny{0.86}}  & 7.55\textsubscript{\tiny{0.35}}  & 31.67\textsubscript{\tiny{13.19}} & 13.76\textsubscript{\tiny{1.67}} & 47.1\textsubscript{\tiny{0.74}} & 43.09\textsubscript{\tiny{0.6}} & 35.26\textsubscript{\tiny{0.33}} & 31.19\textsubscript{\tiny{0.29}} & 45.24\textsubscript{\tiny{2.02}} & 40.14\textsubscript{\tiny{0.79}} & 30.33 & 11.6 & 14 \\
 % \midrule
FixMatch          & 7.72\textsubscript{\tiny{0.33}}  & 7.33\textsubscript{\tiny{0.13}}  & 30.17\textsubscript{\tiny{1.87}} & 11.71\textsubscript{\tiny{1.95}} & 47.61\textsubscript{\tiny{0.83}} & 43.05\textsubscript{\tiny{0.54}} & 33.03\textsubscript{\tiny{0.49}} & 30.51\textsubscript{\tiny{0.53}} & 46.52\textsubscript{\tiny{0.94}} & 40.65\textsubscript{\tiny{0.46}} & 29.83 & 10.2 & 13
 \\

FlexMatch          & 7.82\textsubscript{\tiny{0.77}}  & 7.41\textsubscript{\tiny{0.38}}  & 16.38\textsubscript{\tiny{3.94}} & 12.08\textsubscript{\tiny{0.73}} & 45.73\textsubscript{\tiny{1.6}} & 42.25\textsubscript{\tiny{0.33}} & 35.61\textsubscript{\tiny{1.08}} & 31.13\textsubscript{\tiny{0.18}} & 43.35\textsubscript{\tiny{0.69}} & 40.51\textsubscript{\tiny{0.34}} & 28.23 & 10.0 & 12 \\

SoftMatch          & 7.76\textsubscript{\tiny{0.58}}  & 7.97\textsubscript{\tiny{0.72}}  & 11.9\textsubscript{\tiny{0.27}} & 11.72\textsubscript{\tiny{1.58}} & 45.29\textsubscript{\tiny{0.95}} & 42.21\textsubscript{\tiny{0.2}} & 33.07\textsubscript{\tiny{0.31}} & 30.44\textsubscript{\tiny{0.62}} &44.09\textsubscript{\tiny{0.5}} & 39.76\textsubscript{\tiny{0.13}} & 27.42 & 8.3 & 8
 \\

SimMatch          & 7.93\textsubscript{\tiny{0.55}}  & 7.08\textsubscript{\tiny{0.33}}  & 14.26\textsubscript{\tiny{1.51}} & 12.45\textsubscript{\tiny{1.37}} & 45.91\textsubscript{\tiny{0.95}} & 42.21\textsubscript{\tiny{0.3}} & 33.06\textsubscript{\tiny{0.2}} & 30.16\textsubscript{\tiny{0.21}} &46.12\textsubscript{\tiny{0.48}} & 40.26\textsubscript{\tiny{0.62}} & 27.94 & 8.95 & 10
 \\

CRMatch          & 8.96\textsubscript{\tiny{0.88}}  & 7.16\textsubscript{\tiny{0.09}}  & 12.28\textsubscript{\tiny{1.43}} & 11.08\textsubscript{\tiny{1.24}} & 45.49\textsubscript{\tiny{0.98}} & 43.07\textsubscript{\tiny{0.5}} & 32.51\textsubscript{\tiny{0.4}} & 29.98\textsubscript{\tiny{0.07}} & 45.71\textsubscript{\tiny{0.63}} & 40.62\textsubscript{\tiny{0.28}} & 27.69 & 8.70 & 9
 \\

AdaMatch          & 8.09\textsubscript{\tiny{0.99}}  & 7.11\textsubscript{\tiny{0.2}}  & 11.73\textsubscript{\tiny{0.17}} & 11.22\textsubscript{\tiny{0.95}} & 46.72\textsubscript{\tiny{0.72}} & 42.27\textsubscript{\tiny{0.25}} & 32.75\textsubscript{\tiny{0.35}} & 30.44\textsubscript{\tiny{0.31}} & 45.4\textsubscript{\tiny{0.96}} & 40.16\textsubscript{\tiny{0.49}} & 27.59 & 8.15 & 7
 \\ 
% \midrule

\tiny{VerifyMatch} \textsubscript{\tiny{R}}      & 7.58\textsubscript{\tiny{0.61}} & 
7.38\textsubscript{\tiny{0.43}} & 
11.95\textsubscript{\tiny{0.18}} & 
11.64\textsubscript{\tiny{0.21}} & 
39.97\textsubscript{\tiny{0.29}} & 
40.94\textsubscript{\tiny{0.44}} & 
32.03\textsubscript{\tiny{0.08}} & 
32.14\textsubscript{\tiny{0.71}} & 
37.63\textsubscript{\tiny{0.10}} & 
37.16\textsubscript{\tiny{2.23}} & 25.84 & 6.7 & 6 \\

\midrule
CoTeach \textsubscript{\tiny{R}}    
& \multicolumn{2}{|c|}{9.04\textsubscript{\tiny{2.07}}} 
& \multicolumn{2}{|c|}{11.04\textsubscript{\tiny{0.48}}} 
& \multicolumn{2}{|c|}{39.38\textsubscript{\tiny{0.56}}} 
& \multicolumn{2}{|c|}{31.31\textsubscript{\tiny{0.18}}} 
& \multicolumn{2}{|c|}{36.35\textsubscript{\tiny{1.27}}} 
& 25.42 & 6.7 & 5 \\

% CoTeach \textsubscript{\tiny{M}} & 
% \multicolumn{2}{|c|}{10.44\textsubscript{\tiny{2.95}}} & 
% \multicolumn{2}{|c|}{16.76\textsubscript{\tiny{1.53}}} & 
% \multicolumn{2}{|c|}{39.72\textsubscript{\tiny{0.84}}} & 
% \multicolumn{2}{|c|}{33.67\textsubscript{\tiny{0.34}}} & 
% \multicolumn{2}{|c|}{35.94\textsubscript{\tiny{0.73}}} & 
% 27.31 & 15.30 & 18 \\

% CoTeach \textsubscript{\tiny{L}} & 
% \multicolumn{2}{|c|}{8.30\textsubscript{\tiny{1.75}}} & 
% \multicolumn{2}{|c|}{15.74\textsubscript{\tiny{2.30}}} & 
% \multicolumn{2}{|c|}{43.72\textsubscript{\tiny{1.23}}} & 
% \multicolumn{2}{|c|}{33.45\textsubscript{\tiny{0.52}}} & 
% \multicolumn{2}{|c|}{36.98\textsubscript{\tiny{0.50}}} & 
% 27.64 & 15.50 & 20 \\

% \midrule
\tiny{DivideMix} \textsubscript{\tiny{R}}
& \multicolumn{2}{|c|}{7.67\textsubscript{\tiny{0.17}}} 
& \multicolumn{2}{|c|}{11.05\textsubscript{\tiny{0.14}}} 
& \multicolumn{2}{|c|}{39.34\textsubscript{\tiny{0.44}}} 
& \multicolumn{2}{|c|}{30.83\textsubscript{\tiny{0.66}}} 
& \multicolumn{2}{|c|}{35.34\textsubscript{\tiny{0.55}}} 
& 24.85 & 5.0 & 3 \\

% \tiny{DivideMix} \textsubscript{\tiny{M}}
% & \multicolumn{2}{|c|}{9.98\textsubscript{\tiny{1.90}}} 
% & \multicolumn{2}{|c|}{12.80\textsubscript{\tiny{0.59}}} 
% & \multicolumn{2}{|c|}{38.82\textsubscript{\tiny{0.05}}} 
% & \multicolumn{2}{|c|}{32.04\textsubscript{\tiny{1.57}}} 
% & \multicolumn{2}{|c|}{34.83\textsubscript{\tiny{0.22}}} 
% & 25.69 & 10.70 & 8
%  \\

% \tiny{DivideMix} \textsubscript{\tiny{L}}
% & \multicolumn{2}{|c|}{8.45\textsubscript{\tiny{1.29}}} 
% & \multicolumn{2}{|c|}{13.06\textsubscript{\tiny{0.79}}} 
% & \multicolumn{2}{|c|}{41.95\textsubscript{\tiny{1.73}}} 
% & \multicolumn{2}{|c|}{30.80\textsubscript{\tiny{0.43}}} 
% & \multicolumn{2}{|c|}{35.09\textsubscript{\tiny{0.79}}} 
% & 25.87 & 11.10 & 9
%  \\
% \midrule

Phi-3 
& \multicolumn{2}{|c|}{\textbf{4.78}} 
& \multicolumn{2}{|c|}{12.67} 
& \multicolumn{2}{|c|}{39.32} 
& \multicolumn{2}{|c|}{33.53} 
& \multicolumn{2}{|c|}{34.62} 
& 24.98 & 6.3 & 4 \\
% Mistral 
% & \multicolumn{2}{|c|}{5.08} 
% & \multicolumn{2}{|c|}{14.39} 
% & \multicolumn{2}{|c|}{44.74} 
% & \multicolumn{2}{|c|}{36.27} 
% & \multicolumn{2}{|c|}{38.51} 
% & 27.80 & 14.50 & 15 \\
% Llama-3 
% & \multicolumn{2}{|c|}{\textbf{4.68}} 
% & \multicolumn{2}{|c|}{25.7} 
% & \multicolumn{2}{|c|}{45.67} 
% & \multicolumn{2}{|c|}{39.17} 
% & \multicolumn{2}{|c|}{38.32} 
% & 30.71 & 15.30 & 19
%  \\

\midrule

\tiny{LG-CoTr} \textsubscript{\tiny{B}}       & 7.65\textsubscript{\tiny{0.05}} & 
7.63\textsubscript{\tiny{0.08}} & 
11.36\textsubscript{\tiny{0.06}} & 
10.77\textsubscript{\tiny{0.06}} & 38.12\textsubscript{\tiny{0.06}} & 37.66\textsubscript{\tiny{0.27}} & 29.38\textsubscript{\tiny{0.16}} & \textbf{28.14\textsuperscript{\textdagger}\textsubscript{\tiny{0.23}}} & 33.87\textsubscript{\tiny{0.14}} & 33.52\textsubscript{\tiny{0.12}} & 23.81 & 3.1  & 2 \\

\tiny{LG-CoTr} \textsubscript{\tiny{R}}       & 6.77\textsubscript{\tiny{0.32}} & 
6.58\textsubscript{\tiny{0.15}} & 
11.35\textsubscript{\tiny{0.17}} & 
\textbf{10.41}\textsubscript{\tiny{0.20}} & \textbf{37.15}\textsuperscript{\textdagger}\textsubscript{\tiny{0.19}} & \textbf{37.04\textsuperscript{\textdagger}\textsubscript{\tiny{0.13}}} & \textbf{29.31\textsuperscript{\textdagger}\textsubscript{\tiny{0.11}}} & 28.16\textsubscript{\tiny{0.06}} & \textbf{33.15\textsuperscript{\textdagger}\textsubscript{\tiny{0.27}}} & \textbf{32.93\textsuperscript{\textdagger}\textsubscript{\tiny{0.53}}} & 23.29 & 1.5  & 1 \\

 \bottomrule
\end{tabular}
}
\vspace{-1mm}
\caption{Error rate (\%) and Rank across datasets and label sizes. The first 8 rows are from \citet{wang2022usb}. Subsequent rows show results using pseudo-labels generated by Phi-3 with zero-shot prompting, including our \textbf{LG-CoTr}. Subscripts \textsubscript{R} and \textsubscript{B} indicate RoBERTa and BERT backbones, respectively. Best results are \textbf{bolded}. Superscript \textsuperscript{\textdagger} indicates a statistically significant improvement ($p < 0.05$) over the best baseline using a paired t-test.}

\label{tab:results}
% \vspace{1em}
\vspace{-3mm}

\end{table*}

For LLMs pseudo-labels, we experiment with three open-source LLMs—\textsc{Phi-3-medium-4k}, \textsc{Mistral-7B}, and \textsc{LLaMA-3.1-8B}. We use both zero-shot and few-shot prompting strategies with all three LLMs. Among all settings, \textsc{Phi-3} zero-shot—where \textsc{Phi-3} is also the largest model in terms of parameter count (14B)—achieves the best performance across all test sets; hence, we report the \textsc{Phi-3} zero-shot baseline and the corresponding \textsc{LG-CoTrain} variant that uses pseudo-labels generated by \textsc{Phi-3} zero-shot in the main paper. We show the complete set of prompts for all LLMs in Appendix~\ref{apx:prompts}, with extensive experimental results for both zero-shot and few-shot scenarios being provided in Appendix~\ref{apx:add_results}.

We benchmark our method against a diverse set of SSL baselines using varying amounts of labeled and unlabeled data, following the standardized setup from \citet{wang2022usb}. These baselines include FixMatch \cite{sohn2020fixmatch}, FlexMatch \cite{zhang2021flexmatch}, SoftMatch \cite{chen2023softmatch}, SimMatch \cite{zheng2022simmatch}, CRMatch \cite{fan2023revisiting}, AdaMatch \cite{berthelot2022adamatch}, Dash \cite{xu2021dash}, and UDA \cite{xie2020unsupervised}, along with a supervised-only baseline for reference (as an upper bound), a supervised baseline that uses only the small labeled data, and a supervised baseline that uses the combination of small labeled data and LLM pseudo-labeled data.

To ensure a fair comparison with our dual-classifier architecture, we also include two-network-based methods such as Co-Teaching \cite{NEURIPS2018_a19744e2} and DivideMix \cite{lidividemix}, commonly used in noisy label learning. Additionally, we compare against VerifyMatch \cite{park-caragea-2024-verifymatch}, a recent LLM-guided SSL framework that directly handles noisy pseudo-labels.

\subsection{Results \& Observations}
\label{results}
The comprehensive results of our main experiments, along with comparisons to other baselines, are presented in Table~\ref{tab:results}. Based on these results, we highlight the following key observations.

\vspace{1mm}
\textbf{LG-CoTrain vs. Other SSL Baselines:}  
\textsc{LG-CoTrain} outperforms traditional SSL baselines across all five datasets. While standard methods rely on self-generated pseudo-labels, often amplifying errors, our approach integrates LLM-generated labels and adaptively weights them using training dynamics. Compared to BERT-based results from \citet{wang2022usb}, our BERT-based model (LG-CoTr$_B$) achieves a 3.78\% lower average error than AdaMatch, the strongest of the SSL baselines, demonstrating the effectiveness of our framework.

\textbf{LG-CoTrain vs. VerifyMatch:} VerifyMatch is a semi-supervised learning method that, like our approach, incorporates LLM-generated pseudo-labels into the training process. Thus, it serves as a critical baseline for evaluating the effectiveness of our framework. As shown in Table~\ref{tab:results}, \textsc{LG-CoTrain} outperforms VerifyMatch across all five datasets, with substantial improvements in four of them. On IMDB, our method still achieves a lower error rate, although the margin is relatively small. Overall, \textsc{LG-CoTrain} achieves an average error rate improvement of 2.55\%, further demonstrating the advantage of our method.

% \textbf{LG-CoTrain vs. LLMs:} With the exception of the IMDB dataset, our approach consistently outperforms both the zero-shot and few-shot performance of all three LLMs—Phi-3-medium-4k, Mistral-7B, and LLaMA-3.1-8B—across the test splits of the remaining datasets. Zero-shot results for Phi-3 are reported in Table~\ref{tab:results}, while the zero-shot results for Mistral and LLaMA, along with all few-shot comparisons, are provided in Appendix~\ref{apx:add_results}.

% \textbf{LG-CoTrain vs. LLMs:} With the exception of the IMDB dataset, our approach consistently outperforms both the zero-shot and few-shot performance of all three LLMs—\textsc{Phi-3-medium-4k}, \textsc{Mistral-7B}, and \textsc{LLaMA-3.1-8B}—across the test splits of the remaining datasets. Zero-shot results for Phi-3 are reported in Table~\ref{tab:results}, while the zero-shot results for Mistral and LLaMA, along with all few-shot comparisons, are provided in Appendix~\ref{apx:add_results}. As shown in the Appendix~\ref{apx:add_results}, our best \textsc{LG-CoTrain} variant—ranked highest by the Friedman test—achieves a 1.76\% lower mean error compared to the best-ranked LLM baseline, highlighting its superior overall performance.

% \begin{table*}[htbp]
\begin{table*}[t]
\centering
\scriptsize 

% \caption{Ablation Study Results with Performance Rankings. The best-performing results among all baselines, except \textsc{LG-CoTrain}\textsubscript{oracle}, are highlighted in bold. Friedman Rank and Final Rank show relative performance, with 1 being best.  Superscript \textsuperscript{\textdagger} indicates a statistically significant improvement ($p < 0.05$) over the best baseline using a paired t-test.}

% \label{tab:ablation}

\vspace{-1em}
\begin{tabular}{@{}l|cc|cc|cc|c|c|c@{}}
\toprule
Dataset           & \multicolumn{2}{c}{Amazon Review} & \multicolumn{2}{c}{Yahoo! Answer} & \multicolumn{2}{c|}{Yelp Review} & Mean Error & Friedman & Final \\
\cmidrule(lr){2-3} \cmidrule(lr){4-5} \cmidrule(lr){6-7}
\# Label          & 250             & 1000           & 500             & 2000           & 250             & 1000           & Rate (\%)     & Rank   & Rank \\ 
\midrule

\textsc{LG-CoTrain}\textsubscript{oracle}   & 34.68\textsubscript{\tiny{0.06}} & 34.71\textsubscript{\tiny{0.03}} & 24.27\textsubscript{\tiny{0.04}} & 24.27\textsubscript{\tiny{0.02}} & 30.33\textsubscript{\tiny{0.05}} & 30.52\textsubscript{\tiny{0.24}} & 29.80 & - & - \\

\midrule
\textsc{ST-Random} & 
37.55\textsubscript{\tiny{0.14}} & 
38.08\textsubscript{\tiny{0.21}} & 
30.37\textsubscript{\tiny{0.14}} & 
29.17\textsubscript{\tiny{0.21}} & 
34.14\textsubscript{\tiny{0.29}} & 
34.05\textsubscript{\tiny{0.49}} & 
33.89 & 3.67 & 4 \\

\textsc{FT-Ensembled} & 
41.83\textsubscript{\tiny{0.19}} & 
39.83\textsubscript{\tiny{0.10}} & 
33.60\textsubscript{\tiny{0.15}} & 
32.08\textsubscript{\tiny{0.10}} & 
39.18\textsubscript{\tiny{0.14}} & 
36.74\textsubscript{\tiny{0.08}} & 
37.21 & 5.33 & 5 \\

\textsc{Vanilla-CoTrain} & 
50.37\textsubscript{\tiny{0.53}} & 
49.00\textsubscript{\tiny{0.66}} & 
33.10\textsubscript{\tiny{0.33}} & 
29.29\textsubscript{\tiny{0.82}} & 
43.68\textsubscript{\tiny{0.29}} & 
39.23\textsubscript{\tiny{0.63}} & 
40.78 &  5.67 & 6 \\

\textsc{LG-CoTrain}\textsubscript{SingleSet} & 
37.86\textsubscript{\tiny{0.12}} & 
37.70\textsubscript{\tiny{0.25}} & 
30.03\textsubscript{\tiny{0.12}} & 
28.90\textsubscript{\tiny{0.04}} & 
33.39\textsubscript{\tiny{0.40}} & 
33.01\textsubscript{\tiny{0.35}} & 
33.48 &   2.17 & 2 \\

\textsc{LG-CoTrain}\textsubscript{cc} & 
38.35\textsubscript{\tiny{0.17}} & 
38.06\textsubscript{\tiny{0.21}} & 
30.15\textsubscript{\tiny{0.26}} & 
29.15\textsubscript{\tiny{0.28}} & 
33.83\textsubscript{\tiny{0.26}} & 
33.86\textsubscript{\tiny{0.11}} & 
33.9 & 3.17 & 3 \\

\textsc{LG-CoTrain}       & 
\textbf{37.15}\textsuperscript{\textdagger}\textsubscript{\tiny{0.19}} & \textbf{37.04}\textsuperscript{\textdagger}\textsubscript{\tiny{0.13}} & \textbf{29.31}\textsuperscript{\textdagger}\textsubscript{\tiny{0.11}} & \textbf{28.16}\textsubscript{\tiny{0.06}} & \textbf{33.15}\textsubscript{\tiny{0.27}} & \textbf{32.93}\textsubscript{\tiny{0.53}} & 32.96 & 1 & 1 \\

\bottomrule
\end{tabular}
\caption{Ablation Study for different componenets of \textsc{LG-CoTrain}. The best-performing results among all baselines are highlighted in bold.  Superscript \textsuperscript{\textdagger} indicates a statistically significant improvement ($p < 0.05$) over the best baseline using a paired t-test.}

\label{tab:ablation}
\vspace{-1em}

\end{table*}

\textbf{LG-CoTrain vs. CoTeach and DivideMix:} Our \textsc{LG-CoTrain} method consistently surpasses CoTeach and DivideMix, both of which are designed to mitigate the impact of noisy labels. On average, \textsc{LG-CoTrain} achieves a 1.56\% lower mean error compared to DivideMix and a 2.13\% lower mean error compared to CoTeach, demonstrating the effectiveness of training dynamics-based weighting methods in noisy settings.

% \textbf{LG-CoTrain vs. Supervised Settings:}
% We compare three approaches using the RoBERTa-base backbone: (i) supervised training on a small, accurately labeled subset (\textsc{Sup (lb)}), (ii) supervised training on labeled data augmented with LLM pseudo-labeled data (\textsc{Sup (lb+ps)}), and (iii) \textsc{LG-CoTrain}. From Table~\ref{tab:results}, we can see that adding pseudo-labels to supervised training reduces error by $5.12$ points versus \textsc{Sup (lb)}; \textsc{LG-CoTrain} yields a further $4.28$-point average reduction compared to \textsc{Sup (lb+ps)} (overall $9.40$ points versus \textsc{Sup (lb)}). Gains are largest on the skewed, noisier datasets (Amazon, Yelp, Yahoo), indicating that beyond simple augmentation, \textsc{LG-CoTrain}'s weighting and agreement-aware selection better mitigates pseudo-label noise and class-frequency bias.

\textbf{LG-CoTrain vs. Supervised Settings:}
We compare three baselines to \textsc{LG-CoTrain} (BERT-base): (i) supervised training on the full labeled set (\textsc{Sup (full)}), serving as an empirical upper bound; (ii) supervised training on a small labeled subset (\textsc{Sup (lb)}); and (iii) supervised training on labeled data augmented with LLM pseudo-labels (\textsc{Sup (lb+ps)}). As shown in Table~\ref{tab:results}, as expected, \textsc{Sup (full)} achieves the lowest error (2.22\%), while \textsc{Sup (lb)} performs much worse (33.31\%). In case of (\textsc{Sup (lb+ps)}), adding pseudo-labeled data with the small labeled set helps reduce the mean error by 5.12 points, and \textsc{LG-CoTrain} achieves a further 4.28\% reduction in mean error, totaling a 9.40\% mean error reduction over \textsc{Sup (lb)}. 
% Gains are largest on skewed/noisy datasets (Amazon, Yelp, Yahoo), highlighting that \textsc{LG-CoTrain} better mitigates pseudo-label noise and class imbalance than simple augmentation.

\textbf{LG-CoTrain vs. LLMs:} Among the three LLMs evaluated—\textsc{Phi-3-14B}, \textsc{Mistral-7B}, and \textsc{LLaMA-3.1-8B}—\textsc{Phi-3} achieves the strongest overall performance in the zero-shot setting, while \textsc{LLaMA-3} performs best under few-shot prompting. Despite these strong baselines, our \textsc{LG-CoTrain} approach consistently outperforms both the zero-shot and few-shot results of all three LLMs across the test splits of all datasets, with the sole exception of IMDB (see the comparison with Phi-3 in Table~\ref{tab:results} and the other LLM comparisons in Appendix \ref{apx:add_results}). Notably, our best variant %(\textsc{LG-CoTrain with RoBERTa backbone using LLaMa-3 pseudo labels)} 
achieves a 1.76\% lower mean error than \textsc{Phi-3} zero-shot and a 2.93\% lower mean error than \textsc{LLaMA-3} few-shot.  Comprehensive results with both zero-shot and few-shot approaches are reported in Appendix~\ref{apx:add_results}.

% \cornelia{we did not say that we used both BERT and RoBERTa and why? we do not make it clear that the phy-3 shows the best results and that few shot is the best...}

% \cornelia{also, let's mention the previous paper in related work}

% \cornelia{make a note on error rate}

\section{Analysis}

We conduct our analysis through four complementary investigations: (1) ablation studies examining the contributions of \textsc{LG-CoTrain}'s core components; (2) assessment of \textsc{LG-CoTrain}'s generalizability across natural language understanding (NLU) tasks; (3) analysis of class imbalance and pseudo-label skew effects; and (4) qualitative evaluation of the weighting mechanism's capacity to distinguish clean from noisy samples.

% We perform a comprehensive analysis across four dimensions: (1) \textit{Ablation studies} examining the individual contributions of \textsc{LG-CoTrain}'s core components; (2) \textit{Generalization evaluation} assessing performance on diverse Natural Language Understanding (NLU) tasks; (3) \textit{Robustness analysis} investigating the impact of class imbalance and pseudo-label distribution skew; and (4) \textit{Qualitative assessment} evaluating the weighting mechanism's efficacy in distinguishing clean versus noisy samples.

\subsection{Ablations}

We conducted an ablation study and show the results in Table~\ref{tab:ablation}. The goal of this study is to understand the effectiveness of different components within our framework. Although \textsc{LG-CoTrain} is a cohesive design where multiple components interact synergistically, and it is not always straightforward to isolate the contribution of each part, we aim to analyze their individual impact by systematically removing or modifying specific elements. The results from these controlled variants are reported and analyzed below.

% \textbf{LG-CoTrain vs. Finetuning:}  
% We compare \textsc{LG-CoTrain} against a finetuning-based baseline, where two RoBERTa-base classifiers are first trained on \( \mathcal{D}_{LG} \) and subsequently fine-tuned on \( \mathcal{D}_{l} \). This approach, referred to as \textsc{FT-Ensembled}, results in a 4.25\% increase in mean error rate, highlighting the effectiveness of our weighted co-training mechanism over a purely fine-tuning-based strategy.

\textbf{LG-CoTrain vs. Finetuning:}  
We compare \textsc{LG-CoTrain} with a finetuning-based baseline, where two RoBERTa-base classifiers are trained on the combined dataset \( \mathcal{D}_{LG} \cup \mathcal{D}_{l} \), and their predictions are ensembled during inference. This baseline, referred to as \textsc{FT-Ensembled}, results in a 4.25\% higher mean error rate, underscoring the effectiveness of our weighted co-training strategy—especially in scenarios where the training data includes noisy pseudo-labels generated by LLMs.

% \textbf{LG-CoTrain vs. Self-Training with Random Weighting:}  
% To emphasize the importance of utilizing a pair of backbone classifiers, we compare against \textsc{ST-Random}, where a single classifier is trained with random weights assigned to pseudo-labeled samples using the two metrics from Eq.~\eqref{eq3} and Eq.~\eqref{eq4}. This setup leads to a 0.93\% increase in mean error, reinforcing the benefit of dual classifiers and training dynamics-guided weighting.

\textbf{LG-CoTrain vs. Self-Training with Random Weighting:}  
To highlight the importance of using dual classifiers and training-dynamics-guided weighting, we compare against \textsc{ST-Random}, a self-training baseline where a single classifier is trained using pseudo-labeled samples. For each sample, either \( \lambda_1 \) or \( \lambda_2 \)—as defined in Eq.~\eqref{eq3} and Eq.~\eqref{eq4}—is randomly selected and used as its training weight. This setup results in a 0.93\% higher mean error rate, reinforcing the effectiveness of our dual-classifier design and training dynamics-guided weighting strategy.

\setlength{\tabcolsep}{1.6pt}
\begin{table*}[t]
\centering
\scriptsize

% \vspace{1em}
\begin{tabular}{@{}l|cc|cc|cc|cc|c|c|c@{}}
\toprule
Dataset           & \multicolumn{2}{c}{QQP} & \multicolumn{2}{c}{SWAG} & \multicolumn{2}{c}{MNLI-m}  & \multicolumn{2}{c}{MNLI-mm} & Mean & Fried. & Final \\
\cmidrule(lr){2-3} \cmidrule(lr){4-5} \cmidrule(lr){6-7} \cmidrule(lr){8-9}
\# Label          & 100             & 500           & 200             & 1000           & 150             & 750           & 150             & 750           & Error & Rank   & Rank \\
\midrule

Phi-3 & 
\multicolumn{2}{|c|}{24.66} & 
\multicolumn{2}{|c|}{23.33} & 
\multicolumn{2}{|c|}{19.36} & 
\multicolumn{2}{|c|}{21.53} & 22.22 & 2.38 & 2 \\

CoTeach & 
\multicolumn{2}{|c|}{24.51\textsubscript{\tiny{0.64}}} & 
\multicolumn{2}{|c|}{23.49\textsubscript{\tiny{0.28}}} & 
\multicolumn{2}{|c|}{20.62\textsubscript{\tiny{0.46}}} & 
\multicolumn{2}{|c|}{21.75\textsubscript{\tiny{0.40}}} & 22.59 & 2.88 & 3 \\

\midrule
Vanilla Cotrain & 
32.01\textsubscript{\tiny{0.25}} & 
23.26\textsubscript{\tiny{1.28}} & 
33.57\textsubscript{\tiny{0.38}} & 
27.49\textsubscript{\tiny{0.52}} & 
63.62\textsubscript{\tiny{1.40}} & 
48.72\textsubscript{\tiny{6.44}} & 
59.91\textsubscript{\tiny{3.27}} & 
49.10\textsubscript{\tiny{9.01}} & 42.21 & 3.75 & 4
 \\

LG-CoTrain & 
\textbf{20.22}\textsuperscript{\textdagger}\textsubscript{\tiny{0.77}} & 
\textbf{17.48}\textsuperscript{\textdagger}\textsubscript{\tiny{0.25}} & 
\textbf{22.53}\textsuperscript{\textdagger}\textsubscript{\tiny{0.27}} & 
\textbf{20.91}\textsuperscript{\textdagger}\textsubscript{\tiny{0.30}} & 
\textbf{16.82}\textsuperscript{\textdagger}\textsubscript{\tiny{1.0}} & 
\textbf{15.29}\textsuperscript{\textdagger}\textsubscript{\tiny{0.29}} & 
\textbf{17.32}\textsuperscript{\textdagger}\textsubscript{\tiny{0.94}} & 
\textbf{15.19}\textsuperscript{\textdagger}\textsubscript{\tiny{0.22}} & 18.22 & 1 & 1 \\

\bottomrule
\end{tabular}
\caption{Performance evaluation on the QQP, SWAG, and MNLI development sets. Here, MNLI-m refers to MNLI matched, while MNLI-mm denotes MNLI mismatched. Best results are \textbf{bolded}. \textsuperscript{\textdagger} indicates a statistically significant improvement ($p < 0.05$) over the best baseline using a paired t-test.}
\vspace{-1em}
\label{tab:nlu_results}

\end{table*}
\textbf{LG-CoTrain vs. Vanilla CoTraining:}  
\textsc{Vanilla-CoTrain} relies only on the most confident predictions exchanged between classifier pairs, without leveraging LLM-generated pseudo-labels. This results in a 7.82\% increase in mean error rate, demonstrating the value of incorporating LLM pseudo labels and systematically incorporating them into the training process.

\textbf{LG-CoTrain vs. LG-CoTrain\textsubscript{CC}:}  
\textsc{LG-CoTrain\textsubscript{CC}} is a variant that uses only confidence-based weighting (averaged across iterations), completely ignoring variability. This design reduces divergence between classifiers and weakens performance, leading to a 0.94\% increase in mean error.

\textbf{LG-CoTrain vs. LG-CoTrain (SingleSet):}  
In \textsc{LG-CoTrain (SingleSet)}, both classifiers are trained on the same labeled subset, removing the complementary learning effect gained from disjoint splits. This results in a 0.52\% increase in mean error, indicating that diversity in supervision is important for robust co-training.

\textbf{LG-CoTrain in an Oracle Setting:} 
In Table~\ref{tab:ablation}, \textbf{\textsc{LG-CoTrain}\textsubscript{oracle}}  shows an ideal case where we use only gold-standard (human-annotated) labels instead of LLM-generated pseudo-labels for training. This helps us measure the maximum possible gain from perfect pseudo-labels—resulting in a 3.16\% lower error rate compared to our regular \textsc{LG-CoTrain} that uses LLM-generated labels.

\subsection{Evaluation on Other NLU Tasks}
\label{sec:ood_tasks}

We evaluate \textsc{LG-CoTrain} on a range of challenging natural language understanding (NLU) tasks under different SSL settings. Specifically, we assess its performance on paraphrase detection using the QQP dev set, natural language inference (NLI) using both the MNLI matched and mismatched dev sets, and commonsense reasoning using the SWAG dev set. For each task, we experiment with two SSL settings by varying the number of labeled examples while treating the remaining training data as unlabeled. We compare \textsc{LG-CoTrain} against several baselines, including \textsc{CoTeach}, \textsc{Vanilla CoTraining}, and zero-shot LLM prompting. Across all tasks, \textsc{LG-CoTrain} consistently outperforms the baselines, demonstrating its robustness and effectiveness in NLU tasks under limited supervision. Detailed results are provided in Table~\ref{tab:nlu_results}.

\begin{table}[t]
\centering
\small
\setlength{\tabcolsep}{3pt}
\begin{tabular}{lccc}
\toprule
Dataset & Most (count) & Least (count) & Skew (Most/Least) \\
\midrule
IMDB & 12{,}303 & 10{,}685 & 1.15 \\
AG News & 25{,}545 & 24{,}144 & 1.06 \\
Yahoo & 63{,}659 & 31{,}465 & 2.02 \\
Amazon & 62{,}213 & 30{,}759 & 2.02 \\
Yelp & 67{,}796 & 32{,}420 & 2.09 \\
\bottomrule
\end{tabular}
\caption{\textbf{LLM pseudo-label distribution}. For each dataset we report the most and least frequent pseudo-label counts and the induced skew ratio.}
\label{tab:pseudo_label_dist}
\vspace{-1em}
\end{table}

% \begin{table*}[t]
% \centering
% \small
% \begin{tabular}{@{}l|cc|cc|cc@{}}
% \toprule
% Dataset & \multicolumn{2}{c}{Amazon} & \multicolumn{2}{c}{Yahoo} & \multicolumn{2}{c}{Yelp} \\
% \cmidrule(lr){2-3} \cmidrule(lr){4-5} \cmidrule(lr){6-7}
%  \# Label & 250 & 1000 & 500 & 2000 & 250 & 1000 \\

% \midrule
% VerifyMatch (imb) & 40.71\textsubscript{\tiny{0.76}} & 39.48\textsubscript{\tiny{0.62}} & 32.75\textsubscript{\tiny{0.52}} & 32.26\textsubscript{\tiny{0.48}} & 37.38\textsubscript{\tiny{0.98}} & 37.17\textsubscript{\tiny{0.23}} \\
% LG-CoTrain (imb)  & \textbf{37.77}\textsubscript{\tiny{0.11}} & \textbf{39.01}\textsubscript{\tiny{0.40}} & \textbf{30.71}\textsubscript{\tiny{0.22}} & \textbf{29.46}\textsubscript{\tiny{0.23}} & \textbf{34.92}\textsubscript{\tiny{0.30}} & \textbf{35.73}\textsubscript{\tiny{0.12}} \\
% \bottomrule
% \end{tabular}
% \caption{\textbf{Error rates (\%) under long-tailed imbalance (IR=10).} Both labeled and unlabeled sets follow a long-tailed ground-truth distribution. Lower is better. Means\textsubscript{\tiny{std}} over three runs. Best results are \textbf{bolded}.}
% \label{tab:imbalance}
% \vspace{-1em}
% \end{table*}

\begin{table*}[t]
\centering
\small
% \scriptsize
\begin{tabular}{@{}l|cc|cc|cc|c|c|c@{}}
\toprule
Dataset & \multicolumn{2}{c}{Amazon Review} & \multicolumn{2}{c}{Yahoo Answers} & \multicolumn{2}{c}{Yelp Review} & Mean & Fried. & Final \\
\cmidrule(lr){2-3} \cmidrule(lr){4-5} \cmidrule(lr){6-7}
\# Label & 250 & 1000 & 500 & 2000 & 250 & 1000 & Error & Rank & Rank \\
\midrule

VerifyMatch (imb) & 
40.71\textsubscript{\tiny{0.76}} & 
39.48\textsubscript{\tiny{0.62}} & 
32.75\textsubscript{\tiny{0.52}} & 
32.26\textsubscript{\tiny{0.48}} & 
37.38\textsubscript{\tiny{0.98}} & 
37.17\textsubscript{\tiny{0.23}} & 
36.62 & 2 & 2 \\

LG-CoTrain (imb) & 
\textbf{37.77}\textsubscript{\tiny{0.11}} & 
\textbf{39.01}\textsubscript{\tiny{0.40}} & 
\textbf{30.71}\textsubscript{\tiny{0.22}} & 
\textbf{29.46}\textsubscript{\tiny{0.23}} & 
\textbf{34.92}\textsubscript{\tiny{0.30}} & 
\textbf{35.73}\textsubscript{\tiny{0.12}} & 
34.60 & 1 & 1 \\

\bottomrule
\end{tabular}
\caption{{Error rates (\%) under long-tailed imbalance (IR=10).} Both labeled and unlabeled sets follow a long-tailed ground-truth distribution. Lower is better. Means\textsubscript{\tiny{std}} over three runs. Best results are \textbf{bolded}. Mean error is averaged across all setups. Friedman rank (Fried.) is the average rank across setups, and Final Rank is the overall ranking.}
\label{tab:imbalance}
\vspace{-1em}
\end{table*}

\subsection{Effect of Class Imbalance and Pseudo-Label Skew}
\label{sec:imbalance}

As specified in Appendix~\ref{apx:datasets}, our main experiments use class-balanced labeled and unlabeled training splits. We use the same dataset splits as in \citet{wang2022usb}, which ensures fair and direct comparison with several strong SSL baselines. While the ground-truth label distribution of the training data is balanced in these settings, the \emph{LLM-generated pseudo-labels} over the unlabeled set exhibit varying degrees of skew across datasets: IMDB and AG News maintain near-uniform distributions, while Amazon Review, Yelp Review, and Yahoo Answers exhibit moderate skew with the most frequent class appearing approximately twice as often as the least frequent class (see Table ~\ref{tab:pseudo_label_dist}). This pseudo-label imbalance arises from prediction error by the LLMs. In this scenario—\emph{balanced true labels but imbalanced pseudo-labels}—LG-CoTrain remains robust, as its weighting mechanism down-weights erroneous pseudo-labels and up-weights correct ones (Appendix~\ref{apx:kde_all_datasets}). Thus, the gains in Table~\ref{tab:results} hold even when pseudo-labels are skewed.

% \paragraph{Long-tailed true label distributions.}
% To further stress-test robustness, we also evaluate a second, more challenging scenario in which \emph{both} the labeled and unlabeled sets follow a long-tailed ground-truth distribution with imbalance ratio (IR) 10 (max-to-min class frequency ratio). This induces simultaneous (i) true-class imbalance and (ii) pseudo-label imbalance from the LLM. Table~\ref{tab:imbalance} reports error rates (\%, lower is better) for three datasets (Amazon, Yelp, Yahoo) under two labeled-data budgets. LG-CoTrain consistently outperforms VerifyMatch across all settings, mirroring the trends in Table~\ref{tab:results} and demonstrating robustness to real-world skew.

\paragraph{Long-tailed true label distributions.}
To further stress-test robustness, we evaluate a more challenging scenario where both labeled and unlabeled sets follow a long-tailed ground-truth distribution with an imbalance ratio (IR) of 10 (max-to-min class frequency ratio). This creates simultaneous challenges of (i) true-class imbalance and (ii) pseudo-label noise from LLM predictions. We compare \textsc{LG-CoTrain} against \textsc{VerifyMatch}, a competing method that also utilizes LLM-generated pseudo-labels. The results in Table~\ref{tab:imbalance} show that across three datasets (Amazon, Yelp, Yahoo) and two labeled-data budgets, \textsc{LG-CoTrain} consistently outperforms \textsc{VerifyMatch}. This maintains the performance trends observed in Table~\ref{tab:results} and demonstrates superior robustness to real-world distribution skew.

% \vspace{-1mm}
\subsection{Qualitative Analysis}
\begin{figure}[t]
    \centering
    \includegraphics[width=1.0\linewidth]{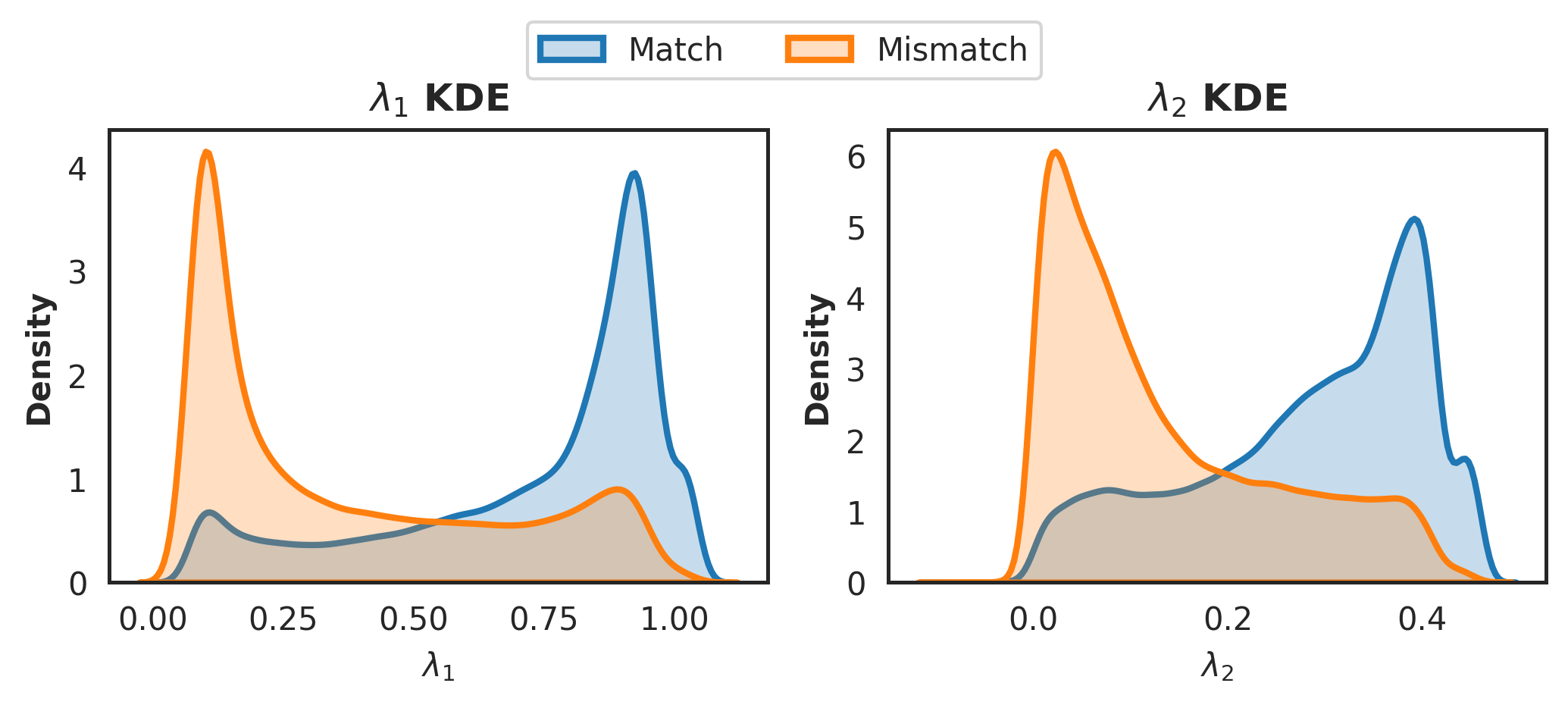} \\

    \caption{KDE plots of $\lambda_1$ and $\lambda_2$ for matched and mismatched samples of Yahoo! Answers dataset.}
    \label{fig:kde_plots_main}
\end{figure}

% \vspace{-1mm}
To better understand the behavior of our sample weighting strategy, we analyze how the learned weights correlate with the correctness of LLM-generated pseudo-labels. In our SSL framework, pseudo-labels remain fixed throughout training, making it essential that correct labels are emphasized while incorrect ones are down-weighted. Figure~\ref{fig:kde_plots_main} presents kernel density estimates (KDE) of the two weighting metrics, $\lambda_1$ and $\lambda_2$, corresponding to the two co-training models trained on \textsc{Yahoo! Answers} dataset. For each plot, we separate samples into two groups: those where the LLM-generated pseudo-label matches the ground truth (Match, shown in blue), and those where it does not (Mismatch, shown in orange).

We observe a clear separation between the two distributions in both $\lambda_1$ and $\lambda_2$. Correctly labeled samples (Match) tend to have higher weights, as indicated by the rightward shift of the blue curves. In contrast, incorrectly labeled samples (Mismatch) exhibit a leftward shift, reflecting lower assigned weights. This confirms that the dynamic weighting effectively distinguishes between high-quality and noisy pseudo-labels, promoting reliable learning signals during co-training. Similar trends are observed across other datasets (see Appendix~\ref{apx:kde_all_datasets}).

% \vspace{-1em}

\section{Conclusion}

In this paper, we introduced \textsc{LG-CoTrain}, a novel semi-supervised learning (SSL) framework that leverages Large Language Model (LLM)-guided pseudo-labeling with weighted error mitigation to enhance co-training. Our approach addresses the key SSL challenge of pseudo-label quality by leveraging LLM-generated labels and refining them with a dynamic weighting mechanism based on training dynamics. Unlike conventional SSL methods that discard low-confidence pseudo-labels, \textsc{LG-CoTrain} utilizes all pseudo-labels, assigning appropriate importance to each to maximize data efficiency. Through extensive experiments on five benchmark datasets, \textsc{LG-CoTrain} consistently outperforms prior SSL baselines, setting new state-of-the-art results. Additionally, we validate its effectiveness across multiple NLU tasks, demonstrating its strong generalization capability. Our findings demonstrate the feasibility of integrating LLMs into SSL frameworks efficiently, opening new avenues for research in semi-supervised NLP. In the future, it would also be interesting to explore approaches where the LLM-generated pseudo-labels are updated over iterations.

% One potential limitation is that we do not evaluate LG-CoTrain in multi-modal setting, which remains an open direction for future research. 

% through a sophisticated interaction between the LLM and the SSL models. 

\section*{Acknowledgement}
This research is supported in part by the NSF IIS award 2107518 and a UIC Discovery Partners Institute (DPI) award. Any opinions, findings, and conclusions expressed here are those of the authors and do not necessarily reflect the views of NSF or DPI. We thank our anonymous reviewers for their constructive feedback, which helped improve the quality of our paper.
% We also thank our reviewers for their insightful feedback and comments.

\section*{Limitations}

One limitation of our approach is the slightly increased computational overhead compared to standard SSL baselines. This is primarily due to the use of a dual-network architecture, which requires training and updating the parameters of two models simultaneously. A detailed comparison of training times is provided in Appendix~\ref{apx:runtime}.

% Bibliography entries for the entire Anthology, followed by custom entries
%\bibliography{anthology,custom}
% Custom bibliography entries only
\bibliography{custom}

% \onecolumn
\appendix

\section{Dataset Details}
\label{apx:datasets}
As previously mentioned, we adopt the same dataset splits as \citet{wang2022usb} to ensure fair comparison with existing benchmarks. Additionally, to evaluate the generalizability of our approach, we incorporate widely-used datasets—MNLI, QQP, and SWAG—from the Hugging Face Datasets Library \cite{lhoest-etal-2021-datasets}. Below, we provide detailed descriptions of each dataset used in our study.

\textbf{IMDB}
The IMDB dataset \cite{maas-etal-2011-learning} is a binary sentiment classification dataset consisting of 25,000 reviews for training and 25,000 for testing. The dataset is balanced, with an equal number of positive and negative reviews in both the training and test sets. For consistency, we rely on the official splits released by \citet{wang2022usb}, which contain 12,500 samples and 1,000 samples per class in the training and validation datasets, respectively. The test set remains unchanged.

% For our work, 12,500 samples and 1,000 samples per class are randomly selected from the training set to form the training and validation datasets. The test set remains unchanged.

\textbf{Amazon Review}
The Amazon Review dataset \cite{mcauley2013hidden} is a sentiment classification dataset with five classes (ratings). Each class contains 600,000 training samples and 130,000 test samples. For consistency, we rely on the official splits released by \citet{wang2022usb}, which contain 50,000 samples and 5,000 samples per class in the training and validation datasets, respectively. The test set remains unchanged.

% For our work, 50,000 samples and 5,000 samples per class are randomly selected from the original training set to form the training and validation datasets. The test set remains unchanged.

\textbf{Yelp Review}
The Yelp Review dataset \cite{yelp_dataset} is a sentiment classification dataset with five classes. It includes 130,000 training samples and 10,000 test samples per class. For consistency, we rely on the official splits released by \citet{wang2022usb}, which contain 50,000 samples and 5,000 samples per class in the training and validation datasets, respectively. The test set remains unchanged.

% For our work, 50,000 samples and 5,000 samples per class are randomly selected from the original training set to form the training and validation datasets. The test set remains unchanged.

\textbf{AG News}
The AG News dataset \cite{NIPS2015_250cf8b5} is a news topic classification dataset with four classes. Each class contains 30,000 training samples and 1,900 test samples. For consistency, we rely on the official splits released by \citet{wang2022usb}, which contain 25,000 samples and 2,500 samples per class in the training and validation datasets, respectively. The test set remains unchanged.

% For our work, 25,000 samples and 2,500 samples per class are randomly selected from the training set to form the training and validation datasets. The test set remains unchanged.

\textbf{Yahoo! Answer}
The Yahoo! Answer dataset \cite{chang2008importance} is a topic classification dataset with ten categories. It includes 140,000 training samples and 6,000 test samples per class. For consistency, we rely on the official splits released by \citet{wang2022usb}, which contain 50,000 samples and 5,000 samples per class in the training and validation datasets, respectively. The test set remains unchanged.

% For our work, 50,000 samples and 5,000 samples per class are randomly selected from the training set to form the training and validation datasets. The test set remains unchanged.

\textbf{QQP}
The Quora Question Pairs (QQP) dataset \cite{qqp} is a binary classification dataset designed to determine whether two questions are semantically equivalent. For our work, we use the training and development splits from the Hugging Face Datasets library, which consist of approximately 363,846 training samples and 40,430 development samples.

\textbf{SWAG Dataset}
The Situations With Adversarial Generations (SWAG) dataset \cite{zellers-etal-2018-swag} is a multiple-choice dataset for commonsense reasoning. For our work, we use the training and development splits from the Hugging Face Datasets library, which consist of 73,546 training samples and 20,006 development samples.

\textbf{MNLI}
The Multi-Genre Natural Language Inference (MNLI) dataset \cite{williams-etal-2018-broad} is a textual entailment dataset with three labels: entailment, contradiction, and neutral. For our work, we use the training and development splits from the Hugging Face Datasets library, which consist of 392,702 training samples and 9815 matched and mismatched developement samples.

% \newpage

% \section{Prompts Used for Pseudo Label Generation}
% \label{sec:prompts}
% In \cref{tab:prompts}, we outline the base prompts utilized for generating pseudo-labels for each dataset. It is important to note that, in addition to these base prompts, we further customize them for each individual LLM by incorporating appropriate system prompts tailored to the model's specific requirements.

\section{Implementation Details} 
\label{apx:implementational_details}
We utilized Llama-3.1-8B \cite{meta2024llama3}, Phi-3-medium-4k \cite{abdin2024phi}, and Mistral-7B-Instruct \cite{mistral2024mistral7b} as our large language models (LLMs) to generate pseudo-labels for unlabeled samples. Throughout this paper, any mention of these models specifically refers to the variants described here. For the backbone models for co-training, we used the base variant of \textsc{RoBERTa} \cite{liu2019roberta} and \textsc{BERT} \cite{devlin-etal-2019-bert}. For all methods, including baselines and our proposed approach, we set the batch size to 32. The models are trained using the Adam optimizer \cite{kingma2014adam} with a learning rate of $2e - 5$. The maximum number of epochs is set to 10 for co-training and 100 for fine-tuning. We apply early stopping with a patience of 5 to prevent overfitting. Each experiment is run three times with different random seeds to ensure the reliability of results.

\section{Prompts Used for Pseudo Label Generation}
\label{apx:prompts}

In Table~\ref{tab:prompts}, we present the base zero-shot prompts used for generating pseudo-labels for each dataset. For few-shot scenarios, we augment these prompts with randomly selected labeled examples from the training set. Specifically, we use 4 examples per class for AG News, 3 for Yahoo! Answers, and 2 for IMDB, Amazon Review, and Yelp Review. 

Additionally, prompts are customized for each LLM by incorporating model-specific system instructions or formatting conventions to ensure compatibility and optimal performance.

\begin{table*}[h]
\centering
\small

% \caption{Example Prompts used to generate pseudo-labels}
% \label{tab:prompts}
\begin{tabular}{@{}l|p{0.75\textwidth}@{}}

\toprule
\textbf{Dataset} & \textbf{Prompt} \\

\midrule
AG News & 
News Article: \texttt{\{news\_article\}}. Based on the content of the news article provided, which of the following categories would it best fit under: 
World, Sports, Business, or Science/Technology? Just select one of these four options. No explanation required. \\

\midrule
IMDB & 
Movie Review: \texttt{\{movie\_review\}}. Based on the content of the movie review provided, determine the category for the movie review: Positive or Negative. 
Select only one of these options. No explanation required. \\

\midrule
Yahoo! Answer & 
Question: \texttt{\{question\}}. Based on the content of the question provided, which of the following categories would it best fit under: 
Society \& Culture, Science \& Mathematics, Health, Education \& Reference, Computers \& Internet, Sports, Business \& Finance, Entertainment \& Music, Family \& Relationships, or Politics \& Government? 
Just select one of these ten options. No explanation required. \\

\midrule
Amazon Reviews & 
Review Text: \texttt{\{review\_text\}}. Based on the content of the review text provided, determine the rating for the product review: (1 star, 2 stars, 3 stars, 4 stars, or 5 stars). 
Just select one of these five options. No explanation required. \\

\midrule
Yelp Reviews & 
Review Text: \texttt{\{review\_text\}}. Based on the content of the review text provided, determine the rating for the restaurant review: (1 star, 2 stars, 3 stars, 4 stars, or 5 stars). 
Just select one of these five options. No explanation required. \\

\midrule
QQP & 
Question 1: \texttt{\{question1\}}. Question 2: \texttt{\{question2\}}. 
Based on the context, phrasing, and intent of the two questions provided, determine if the two questions are semantically equivalent (duplicates). 
Select only one of these options: duplicate or not duplicate. No explanation required. \\

\midrule
SWAG & 
Startphrase: \texttt{\{startphrase\}}. Ending 0: \texttt{\{ending0\}}. 
Ending 1: \texttt{\{ending1\}}. Ending 2: \texttt{\{ending2\}}. Ending 3: \texttt{\{ending3\}}. 
Based on the startphrase and potential endings provided, determine the most likely ending to the startphrase. 
Respond with only one of these options: 0, 1, 2, or 3. No explanation required. \\

\midrule
MNLI & 
Premise: \texttt{\{premise\}}. Hypothesis: \texttt{\{hypothesis\}}. 
Based on the premise and hypothesis provided, determine the most likely relationship between the two. 
Respond with only one of these options: entailment, contradiction, or neutral. No explanation required. \\

\bottomrule
\end{tabular}
\caption{Example Prompts used to generate pseudo-labels}
\label{tab:prompts}
\end{table*}

% \newpage

% \section{Additional Results}
% \label{apx:add_results}
% In this section, we present extended experimental results that include additional LLM pseudo-labelers: \textbf{LLaMA-3}, \textbf{Mistral}, and \textbf{Phi-3}. We evaluate their performance under both \textbf{zero-shot} and \textbf{few-shot} prompting settings. Table~\ref{tab:more_results} complements the main paper by offering a comprehensive comparison across datasets and label sizes, using error rate and ranking metrics. One interesting fact is that, in some cases, few-shot in context learning underperfomrs than zero-shot settings, we attribute this to the fact that the few-shot examples used in our prompts were randomly sampled from the training set and as prior work \cite{zhang-etal-2022-active}, the selection of few-shot exemplars plays a critical role in in-context learning performance. So selecting the randomly selected examples causes the performance to drop sometimes.

\section{Additional Results}
\label{apx:add_results}

This section presents extended experimental results incorporating three additional LLM pseudo-labelers: \textbf{LLaMA-3}, \textbf{Mistral}, and \textbf{Phi-3}. We evaluate these models under both \textbf{zero-shot} and \textbf{few-shot} prompting configurations. Table~\ref{tab:more_results} provides a comprehensive comparison across datasets and label sizes, reporting both error rates and ranking metrics to complement the main paper's findings.

Notably, we observe that few-shot in-context learning occasionally underperforms compared to zero-shot settings. We attribute this phenomenon to the random sampling of few-shot examples from the training set, as prior work \cite{zhang-etal-2022-active} demonstrates that exemplar selection critically impacts in-context learning performance. The suboptimal performance in some few-shot scenarios suggests that randomly selected examples may introduce noise or bias that degrades pseudo-labeling quality.

% \setlength{\tabcolsep}{1.6pt} % global spacing between columns
% \begin{table*}[h]
% \centering
% \renewcommand{\arraystretch}{1.1} %global vertical spacing between rows
% \scriptsize 

% \caption{Error rate (\%) and Rank across datasets and label sizes. Subscripts \textsubscript{P}, \textsubscript{M}, and \textsubscript{L} denote the LLMs used: Phi-3, Mistral, and LLaMA-3, respectively. Subscripts \textsubscript{FS} and \textsubscript{ZS} indicate few-shot and zero-shot prompting settings. Best results are \textbf{bolded}. \textsuperscript{\textdagger} indicates a statistically significant improvement ($p < 0.05$) over the best baseline using a paired t-test. Grayed rows are also reported in the main paper.}

\setlength{\tabcolsep}{1.6pt} % global spacing between columns
\begin{table*}[h]
\centering
\renewcommand{\arraystretch}{1.1} %global vertical spacing between rows
\scriptsize 
\vspace{1em}

% \caption{Error rate (\%) and Rank across datasets and label sizes. Subscripts \textsubscript{P}, \textsubscript{M}, and \textsubscript{L} denote the LLMs used: Phi-3, Mistral, and LLaMA-3, respectively. Subscripts \textsubscript{FS} and \textsubscript{ZS} indicate few-shot and zero-shot prompting settings. Best results are \textbf{bolded}. \textsuperscript{\textdagger} indicates a statistically significant improvement ($p < 0.05$) over the best baseline using a paired t-test. Grayed rows are also reported in the main paper.}
% \label{tab:more_results}

\resizebox{0.96\textwidth}{!}{
\begin{tabular}{l|cc|cc|cc|cc|cc|c|c|c}
\toprule
Dataset           & \multicolumn{2}{c|}{IMDB}         & \multicolumn{2}{c|}{AG News}      & \multicolumn{2}{c|}{Amazon Review} & \multicolumn{2}{c|}{Yahoo! Answer} & \multicolumn{2}{c|}{Yelp Review}  & Mean & Fried. & Final \\ 
% \midrule
% \cmidrule(lr){1-1} \cmidrule(lr){2-3} \cmidrule(lr){4-5} \cmidrule(lr){6-7} \cmidrule(lr){8-9} \cmidrule(lr){10-11} \cmidrule(lr){12-14}
\# Label          & 20             & 40             & 40           & 200           & 250             & 1000           & 500             & 2000           & 250             & 1000           & Err.            & rank             & rank           \\ \midrule

LG-CoTr \textsubscript{\tiny{P-FS}}       & 7.11\textsubscript{\tiny{0.15}} & 
6.99\textsubscript{\tiny{0.20}} & 
13.22\textsubscript{\tiny{1.24}} & 
11.29\textsubscript{\tiny{0.50}} & 
\textbf{36.76}\textsuperscript{\textdagger}\textsubscript{\tiny{0.05}} & 
\textbf{36.70}\textsuperscript{\textdagger}\textsubscript{\tiny{0.09}} & 
29.46\textsubscript{\tiny{0.07}} & 
28.62\textsubscript{\tiny{0.09}} & 
33.46\textsubscript{\tiny{0.52}} & 
32.85\textsubscript{\tiny{0.61}}
 & 23.65 & 5.55 & 4 \\

LG-CoTr \textsubscript{\tiny{M-FS}}       & 7.11\textsubscript{\tiny{0.06}} & 
6.83\textsubscript{\tiny{0.11}} & 
11.96\textsubscript{\tiny{0.80}} & 
11.35\textsubscript{\tiny{0.23}} & 
37.53\textsubscript{\tiny{0.23}} & 
36.85\textsubscript{\tiny{0.34}} & 
30.22\textsubscript{\tiny{0.13}} & 
28.75\textsubscript{\tiny{0.15}} & 
33.80\textsubscript{\tiny{0.14}} & 
33.04\textsubscript{\tiny{0.29}}

 & 23.74 & 6.35 & 6 \\

LG-CoTr \textsubscript{\tiny{L-FS}}       & 6.63\textsubscript{\tiny{0.35}} & 
6.43\textsubscript{\tiny{0.32}} & 
11.12\textsubscript{\tiny{0.11}} & 
10.70\textsubscript{\tiny{0.27}} & 
37.47\textsubscript{\tiny{0.24}} & 
36.71\textsubscript{\tiny{0.18}} & 
\textbf{28.67}\textsuperscript{\textdagger}\textsubscript{\tiny{0.19}} & 
\textbf{27.97}\textsuperscript{\textdagger}\textsubscript{\tiny{0.14}} & 
33.60\textsubscript{\tiny{0.13}} & 
32.91\textsubscript{\tiny{0.17}}
 & 23.22 & 3.10 & 1 \\

\midrule

\mycc LG-CoTr \textsubscript{\tiny{P-ZS}}       & \mycc 6.77\textsubscript{\tiny{0.32}} & 
\mycc 6.58\textsubscript{\tiny{0.15}} & 
\mycc 11.35\textsubscript{\tiny{0.17}} & 
\mycc 10.41\textsubscript{\tiny{0.20}} & \mycc 37.15\textsubscript{\tiny{0.19}} & \mycc 37.04\textsubscript{\tiny{0.13}} & \mycc 29.31\textsubscript{\tiny{0.11}} & \mycc 28.16\textsubscript{\tiny{0.06}} & \mycc \textbf{33.15}\textsuperscript{\textdagger}\textsubscript{\tiny{0.27}} & \mycc 32.93\textsubscript{\tiny{0.53}} & 23.29 & 3.70 & 2 \\

LG-CoTr \textsubscript{\tiny{M-ZS}} & 6.78\textsubscript{\tiny{0.23}} & 
6.94\textsubscript{\tiny{0.04}} & 11.19\textsubscript{\tiny{0.12}} & \textbf{10.01}\textsubscript{\tiny{0.30}} & 37.91\textsubscript{\tiny{0.33}} & 
37.21\textsubscript{\tiny{0.10}} & 
30.20\textsubscript{\tiny{0.09}} & 
28.96\textsubscript{\tiny{0.20}} & 
33.68\textsubscript{\tiny{0.27}} & 
\textbf{32.76}\textsuperscript{\textdagger}\textsubscript{\tiny{0.07}} & 23.56 & 5.00 & 3 \\

LG-CoTr \textsubscript{\tiny{L-ZS}} & 6.81\textsubscript{\tiny{0.27}} & 6.68\textsubscript{\tiny{0.26}} & 
12.29\textsubscript{\tiny{0.52}} & 
10.77\textsubscript{\tiny{0.81}} & 
38.49\textsubscript{\tiny{0.07}} & 
37.53\textsubscript{\tiny{0.23}} & 
29.18\textsubscript{\tiny{0.16}} & 
28.29\textsubscript{\tiny{0.06}} & 
33.85\textsubscript{\tiny{0.48}} & 
33.28\textsubscript{\tiny{0.12}} & 23.71 & 6.00 & 5 \\

% LG-CoTr \textsubscript{\tiny{R}} & 24.41\textsubscript{\tiny{1.48}} & 
% 13.48\textsubscript{\tiny{0.32}} & 61.22\textsubscript{\tiny{15.90}} & 11.80\textsubscript{\tiny{0.94}} & 57.85\textsubscript{\tiny{0.24}} & 41.80\textsubscript{\tiny{0.18}} & 37.62\textsubscript{\tiny{1.05}} & 29.76\textsubscript{\tiny{0.14}} & 51.11\textsubscript{\tiny{3.09}} & 38.40\textsubscript{\tiny{0.19}} & 36.65 & 18 & 18 \\

\midrule

\mycc \tiny{VerifyMatch} \textsubscript{\tiny{P-ZS}}       & \mycc 7.58\textsubscript{\tiny{0.61}} & 
\mycc 7.38\textsubscript{\tiny{0.43}} & 
\mycc 11.95\textsubscript{\tiny{0.18}} & 
\mycc 11.64\textsubscript{\tiny{0.21}} & 
\mycc 39.97\textsubscript{\tiny{0.29}} & 
\mycc 40.94\textsubscript{\tiny{0.44}} & 
\mycc 32.03\textsubscript{\tiny{0.08}} & 
\mycc 32.14\textsubscript{\tiny{0.71}} & 
\mycc 37.63\textsubscript{\tiny{0.10}} & 
\mycc 37.16\textsubscript{\tiny{2.23}} & 25.84 & 12.30 & 11 \\

\tiny{VerifyMatch} \textsubscript{\tiny{M-ZS}} & 
8.29\textsubscript{\tiny{0.69}} & 
8.00\textsubscript{\tiny{0.33}} & 
13.33\textsubscript{\tiny{1.13}} & 
14.22\textsubscript{\tiny{1.93}} & 
40.41\textsubscript{\tiny{0.77}} & 
40.03\textsubscript{\tiny{1.24}} & 
34.52\textsubscript{\tiny{0.51}} & 
34.17\textsubscript{\tiny{0.36}} & 
35.45\textsubscript{\tiny{0.53}} & 
35.74\textsubscript{\tiny{0.19}} & 
26.42 & 14.40 & 15 \\

\tiny{VerifyMatch} \textsubscript{\tiny{L-ZS}} & 
7.57\textsubscript{\tiny{0.66}} & 
7.90\textsubscript{\tiny{0.85}} & 
15.27\textsubscript{\tiny{1.55}} & 
14.31\textsubscript{\tiny{1.36}} & 
45.88\textsubscript{\tiny{1.26}} & 
43.86\textsubscript{\tiny{0.74}} & 
33.78\textsubscript{\tiny{0.25}} & 
33.43\textsubscript{\tiny{0.51}} & 
37.79\textsubscript{\tiny{0.44}} & 
39.58\textsubscript{\tiny{1.23}} & 
27.94 & 16.30 & 20 \\

\midrule

\mycc CoTeach \textsubscript{\tiny{P-ZS}}       
& \multicolumn{2}{|c|}{\mycc 9.04\textsubscript{\tiny{2.07}}} 
& \multicolumn{2}{|c|}{\mycc 11.04\textsubscript{\tiny{0.48}}} 
& \multicolumn{2}{|c|}{\mycc 39.38\textsubscript{\tiny{0.56}}} 
& \multicolumn{2}{|c|}{\mycc 31.31\textsubscript{\tiny{0.18}}} 
& \multicolumn{2}{|c|}{\mycc 36.35\textsubscript{\tiny{1.27}}} 
& 25.42 & 11.0 & 9 \\

CoTeach \textsubscript{\tiny{M-ZS}} & 
\multicolumn{2}{|c|}{10.44\textsubscript{\tiny{2.95}}} & 
\multicolumn{2}{|c|}{16.76\textsubscript{\tiny{1.53}}} & 
\multicolumn{2}{|c|}{39.72\textsubscript{\tiny{0.84}}} & 
\multicolumn{2}{|c|}{33.67\textsubscript{\tiny{0.34}}} & 
\multicolumn{2}{|c|}{35.94\textsubscript{\tiny{0.73}}} & 
27.31 & 15.50 & 17 \\

CoTeach \textsubscript{\tiny{L-ZS}} & 
\multicolumn{2}{|c|}{8.30\textsubscript{\tiny{1.75}}} & 
\multicolumn{2}{|c|}{15.74\textsubscript{\tiny{2.30}}} & 
\multicolumn{2}{|c|}{43.72\textsubscript{\tiny{1.23}}} & 
\multicolumn{2}{|c|}{33.45\textsubscript{\tiny{0.52}}} & 
\multicolumn{2}{|c|}{36.98\textsubscript{\tiny{0.50}}} & 
27.64 & 15.70 & 18 \\

\midrule

\mycc \tiny{DivideMix} \textsubscript{\tiny{P-ZS}}
& \multicolumn{2}{|c|}{\mycc 7.67\textsubscript{\tiny{0.17}}} 
& \multicolumn{2}{|c|}{\mycc 11.05\textsubscript{\tiny{0.14}}} 
& \multicolumn{2}{|c|}{\mycc 39.34\textsubscript{\tiny{0.44}}} 
& \multicolumn{2}{|c|}{\mycc 30.83\textsubscript{\tiny{0.66}}} 
& \multicolumn{2}{|c|}{\mycc 35.34\textsubscript{\tiny{0.55}}} 
& 24.85 & 9.30 & 8 \\

\tiny{DivideMix} \textsubscript{\tiny{M-ZS}}
& \multicolumn{2}{|c|}{9.98\textsubscript{\tiny{1.90}}} 
& \multicolumn{2}{|c|}{12.80\textsubscript{\tiny{0.59}}} 
& \multicolumn{2}{|c|}{38.82\textsubscript{\tiny{0.05}}} 
& \multicolumn{2}{|c|}{31.99\textsubscript{\tiny{0.99}}} 
& \multicolumn{2}{|c|}{34.83\textsubscript{\tiny{0.22}}} 
& 25.68 & 11.30 & 10
 \\

\tiny{DivideMix} \textsubscript{\tiny{L-ZS}}
& \multicolumn{2}{|c|}{8.45\textsubscript{\tiny{1.29}}} 
& \multicolumn{2}{|c|}{13.06\textsubscript{\tiny{0.79}}} 
& \multicolumn{2}{|c|}{41.95\textsubscript{\tiny{1.73}}} 
& \multicolumn{2}{|c|}{30.81\textsubscript{\tiny{0.31}}} 
& \multicolumn{2}{|c|}{35.09\textsubscript{\tiny{0.79}}} 
& 25.87 & 12.50 & 12
 \\
\midrule

Phi-3 \textsubscript{\tiny{FS}}
& \multicolumn{2}{|c|}{11.84} 
& \multicolumn{2}{|c|}{18.07} 
& \multicolumn{2}{|c|}{37.5} 
& \multicolumn{2}{|c|}{38.52} 
& \multicolumn{2}{|c|}{34.14} 
& 28.01 & 14.20 & 14
 \\
Mistral \textsubscript{\tiny{FS}}
& \multicolumn{2}{|c|}{5.79} 
& \multicolumn{2}{|c|}{19.62} 
& \multicolumn{2}{|c|}{41.67} 
& \multicolumn{2}{|c|}{39.45} 
& \multicolumn{2}{|c|}{39.18} 
& 29.14 & 16.30 & 21 \\
Llama-3 \textsubscript{\tiny{FS}}
& \multicolumn{2}{|c|}{6.37} 
& \multicolumn{2}{|c|}{13.09} 
& \multicolumn{2}{|c|}{40.14} 
& \multicolumn{2}{|c|}{33.96} 
& \multicolumn{2}{|c|}{37.2} 
& 26.15 & 12.80 & 13
 \\

 \midrule

% \tiny{Llama-3} \textsubscript{\tiny{zeroshot}}
\mycc Phi-3 \textsubscript{\tiny{ZS}}
& \multicolumn{2}{|c|}{\mycc 4.78} 
& \multicolumn{2}{|c|}{\mycc 12.67} 
& \multicolumn{2}{|c|}{\mycc 39.32} 
& \multicolumn{2}{|c|}{\mycc 33.53} 
& \multicolumn{2}{|c|}{\mycc 34.62} 
& 24.98 & 8.40 & 7 \\
Mistral \textsubscript{\tiny{ZS}}
& \multicolumn{2}{|c|}{5.08} 
& \multicolumn{2}{|c|}{14.39} 
& \multicolumn{2}{|c|}{44.74} 
& \multicolumn{2}{|c|}{36.27} 
& \multicolumn{2}{|c|}{38.51} 
& 27.80 & 15.10 & 16 \\
Llama-3 \textsubscript{\tiny{ZS}}
& \multicolumn{2}{|c|}{\textbf{4.68}} 
& \multicolumn{2}{|c|}{25.7} 
& \multicolumn{2}{|c|}{45.67} 
& \multicolumn{2}{|c|}{39.17} 
& \multicolumn{2}{|c|}{38.32} 
& 30.71 & 16.20 & 19
 \\

 \bottomrule
\end{tabular}
}

\caption{Error rate (\%) and Rank across datasets and label sizes. Subscripts \textsubscript{P}, \textsubscript{M}, and \textsubscript{L} denote the LLMs used: Phi-3, Mistral, and LLaMA-3, respectively. Subscripts \textsubscript{FS} and \textsubscript{ZS} indicate few-shot and zero-shot prompting settings. Best results are \textbf{bolded}. \textsuperscript{\textdagger} indicates a statistically significant improvement ($p < 0.05$) over the best baseline using a paired t-test. Grayed rows are also reported in the main paper.}
\label{tab:more_results}
\vspace{1em}
\end{table*}

\section{Training Cost}
\label{apx:runtime}
To ensure a fair comparison across all methods, we standardized the training configuration for each experiment. Table~\ref{tab:experimental_setup} summarizes the compute budget used for each method and dataset, including the number of training hours, training settings per run, and the number of random seeds. The total GPU hours reported assume a single GPU per run. All experiments were conducted using NVIDIA V100 GPUs.

\setlength{\tabcolsep}{10pt}
\begin{table*}[t]
\centering
\small

\begin{tabular}{ll>{\centering\arraybackslash}p{3cm}l}
\toprule
\textbf{Method} & \textbf{Dataset} & \textbf{Hours × Settings × Seeds} & \textbf{Total GPU Hours} \\
\midrule
\multirow{5}{*}{USB SSL} 
    & IMDB & $8 \times 2 \times 3$ & \multirow{5}{*}{\makecell{222 GPU Hours \\ (9 GPU Days)}} \\
    & AG News & $6 \times 2 \times 3$ & \\
    & Amazon Review & $8 \times 2 \times 3$ & \\
    & Yahoo! Answer & $7 \times 2 \times 3$ & \\
    & Yelp Review & $8 \times 2 \times 3$ & \\
\midrule
\multirow{5}{*}{VerifyMatch} 
    & IMDB & $1 \times 2 \times 3$ & \multirow{5}{*}{\makecell{195 GPU Hours \\ (8 GPU Days)}} \\
    & AG News & $2.5 \times 2 \times 3$ & \\
    & Amazon Review & $8 \times 2 \times 3$ & \\
    & Yahoo! Answer & $13 \times 2 \times 3$ & \\
    & Yelp Review & $8 \times 2 \times 3$ & \\
\midrule
\multirow{5}{*}{LG-CoTrain} 
    & IMDB & $1 \times 2 \times 3$ & \multirow{5}{*}{\makecell{294 GPU Hours \\ (12 GPU Days)}} \\
    & AG News & $4 \times 2 \times 3$ & \\
    & Amazon Review & $12 \times 2 \times 3$ & \\
    & Yahoo! Answer & $20 \times 2 \times 3$ & \\
    & Yelp Review & $12 \times 2 \times 3$ & \\
\bottomrule
\end{tabular}
\caption{Comparison of GPU Budget and Training Setup for Each Method and Dataset}
\label{tab:experimental_setup}
\end{table*}

\subsection{Fairness in Parameter and Inference Cost}
\label{apx:self-ensemble}
Our proposed framework requires training two models jointly and performing inference with both at test time. To ensure fair comparisons, we further evaluated \textit{self-ensembled} baselines, where each baseline (e.g., AdaMatch, SoftMatch) is trained twice independently and their predictions are ensembled at inference. This matches our framework in terms of parameter count and test-time computation.

The results, shown in Table~\ref{tab:self_ensemble}, indicate that our method consistently outperforms these self-ensembled variants, demonstrating that our co-training design provides benefits beyond simply increasing model size or applying post-hoc ensembling.

\begin{table*}[t]
\centering
\small
\begin{tabular}{@{}l|cc|cc|cc@{}}
\toprule
Dataset & \multicolumn{2}{c}{Amazon} & \multicolumn{2}{c}{Yahoo} & \multicolumn{2}{c}{Yelp} \\
\cmidrule(lr){2-3} \cmidrule(lr){4-5} \cmidrule(lr){6-7}
\# Label & 250 & 1000 & 500 & 2000 & 250 & 1000 \\
\midrule
AdaMatch-En  & 46.24\textsubscript{\tiny{0.52}} & 41.27\textsubscript{\tiny{0.09}} & 30.53\textsubscript{\tiny{0.40}} & 28.39\textsubscript{\tiny{0.24}} & 44.31\textsubscript{\tiny{0.43}} & 37.96\textsubscript{\tiny{0.19}} \\
SoftMatch-En & 42.97\textsubscript{\tiny{0.35}} & 41.15\textsubscript{\tiny{0.29}} & 30.24\textsubscript{\tiny{0.17}} & 28.48\textsubscript{\tiny{0.12}} & 41.71\textsubscript{\tiny{0.35}} & 37.44\textsubscript{\tiny{0.18}} \\
Ours         & \textbf{38.12}\textsubscript{\tiny{0.06}} & \textbf{37.66}\textsubscript{\tiny{0.27}} & \textbf{29.38}\textsubscript{\tiny{0.16}} & \textbf{28.14}\textsubscript{\tiny{0.23}} & \textbf{33.87}\textsubscript{\tiny{0.14}} & \textbf{33.52}\textsubscript{\tiny{0.12}} \\
\bottomrule
\end{tabular}
\caption{\textbf{Error rates (\%) of ensemble variants.} Comparison between AdaMatch-Ensembled, SoftMatch-Ensembled, and our method on three datasets. Lower is better. Means\textsubscript{\tiny{std}} over three runs. Best results are \textbf{bolded}.}
\label{tab:self_ensemble}
\vspace{-1em}
\end{table*}

\section{Distribution of Pseudo-Label Weights}

\label{apx:kde_all_datasets}

\begin{figure*}[t]
    \centering

    \begin{subfigure}{0.5\linewidth}
        \centering
        \includegraphics[width=\linewidth]{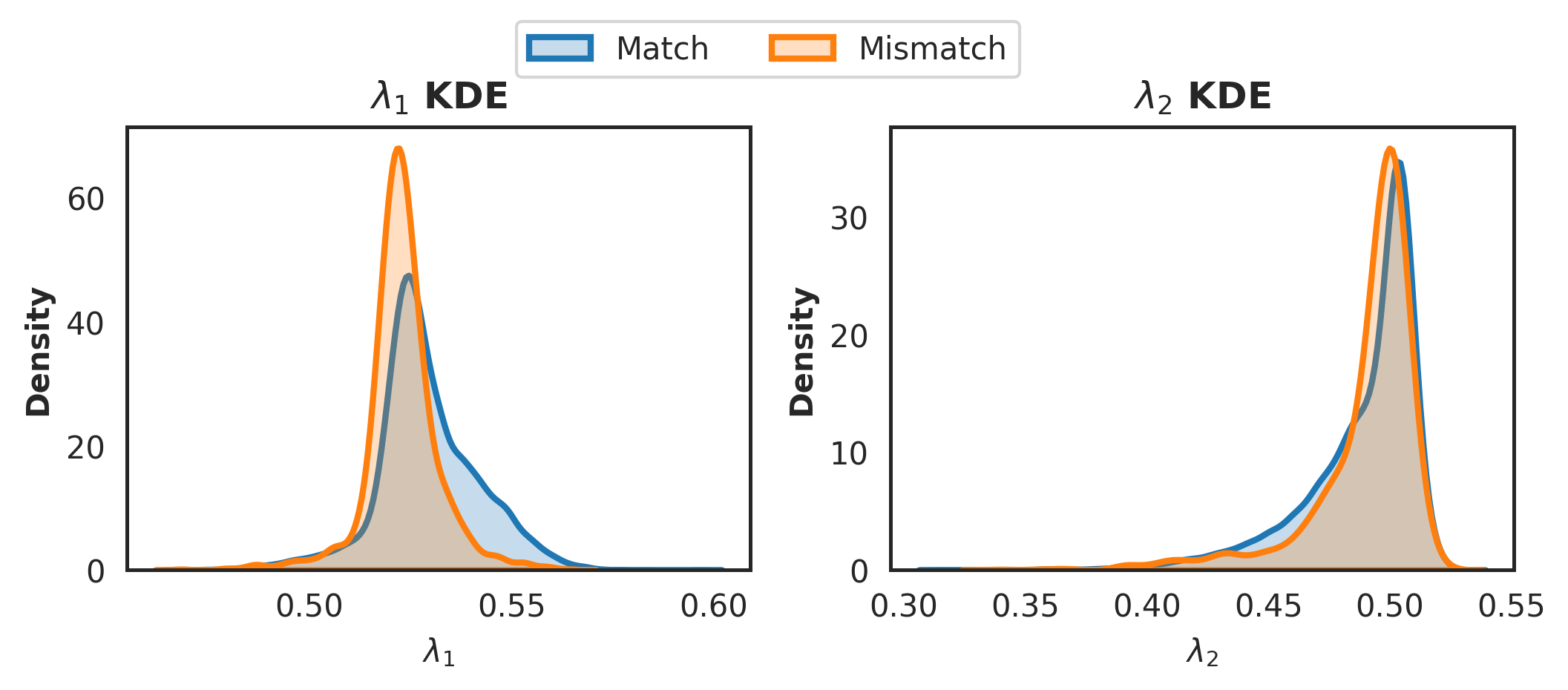}
        \caption{\textsc{IMDB}}
    \end{subfigure}

    \begin{subfigure}{0.5\linewidth}
        \centering
        \includegraphics[width=\linewidth]{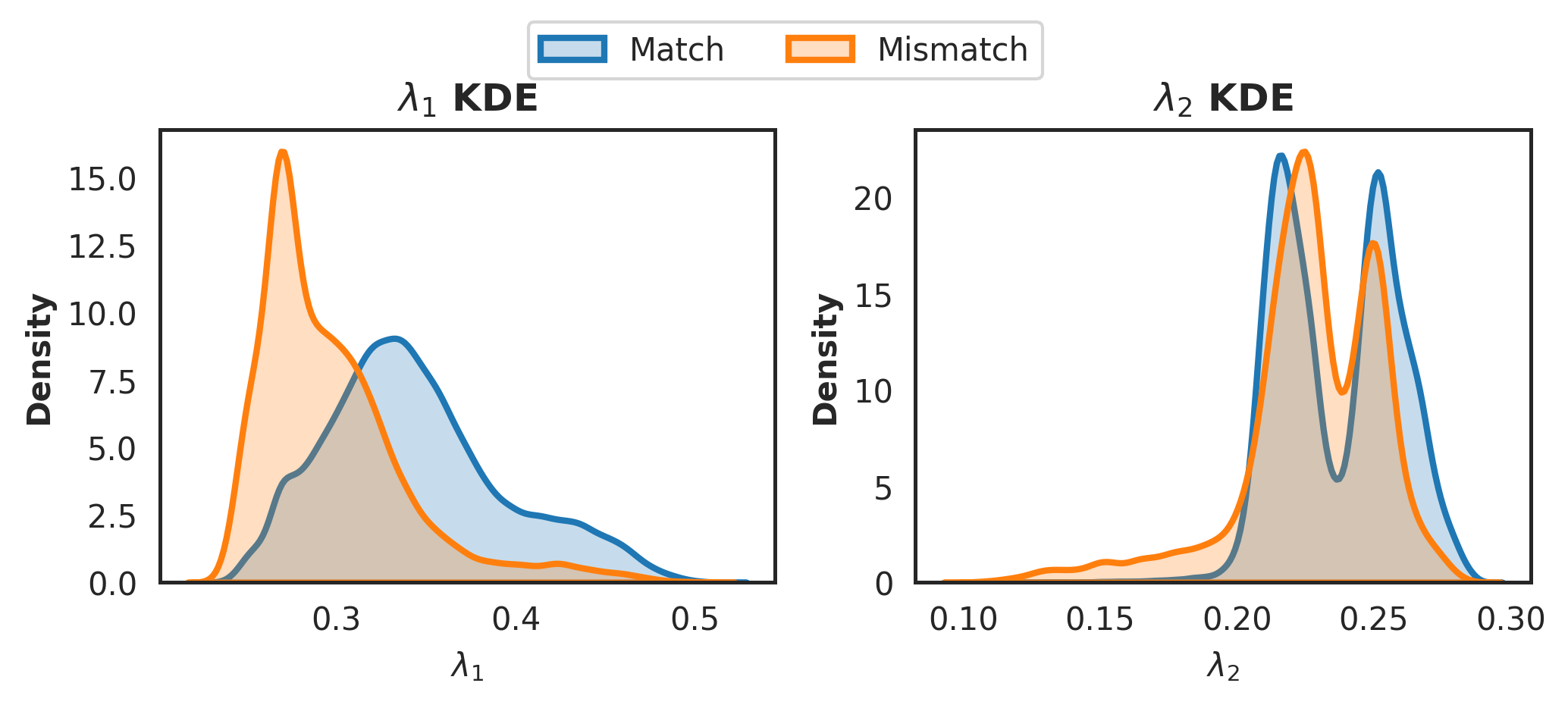}
        \caption{\textsc{AG News}}
    \end{subfigure}

    \begin{subfigure}{0.5\linewidth}
        \centering
        \includegraphics[width=\linewidth]{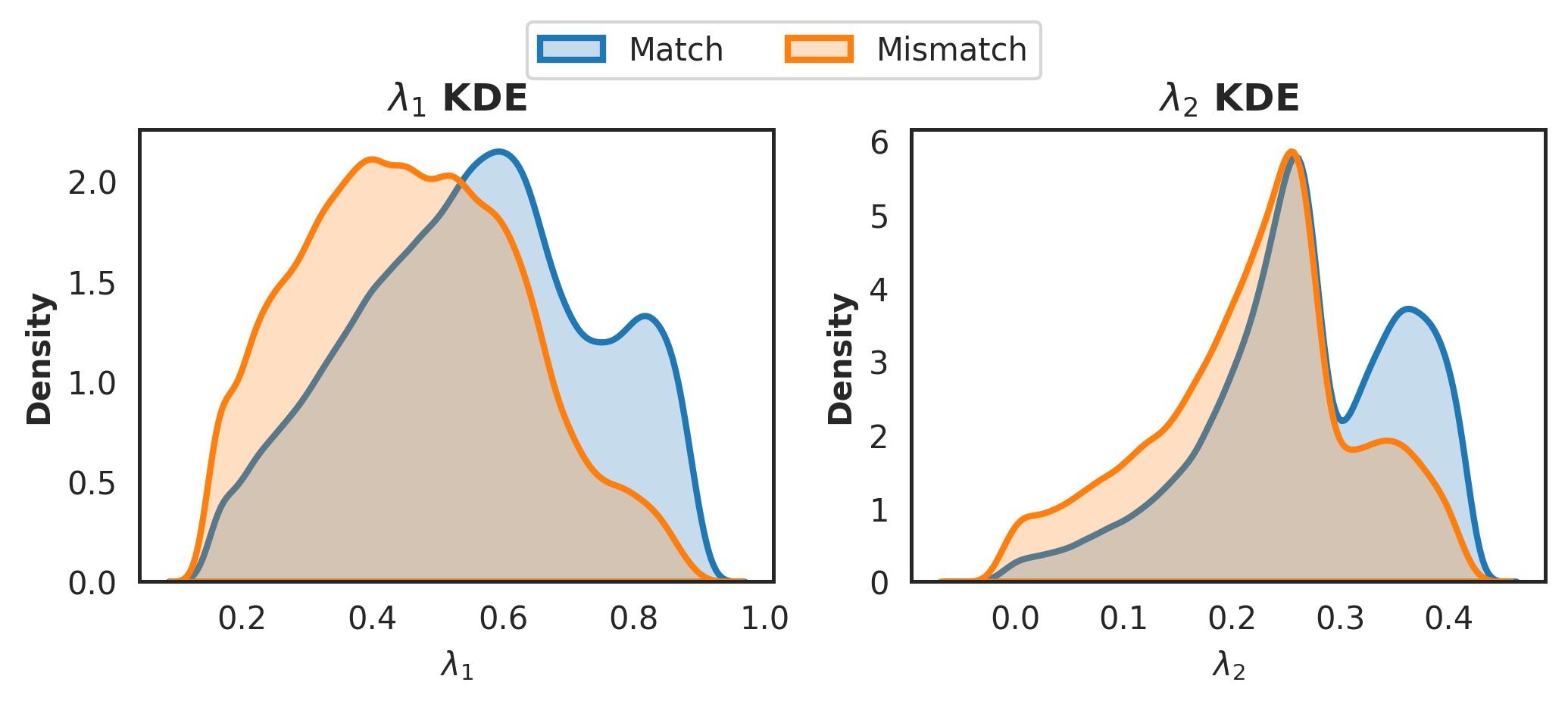}
        \caption{\textsc{Amazon Review}}
    \end{subfigure}

    \begin{subfigure}{0.5\linewidth}
        \centering
        \includegraphics[width=\linewidth]{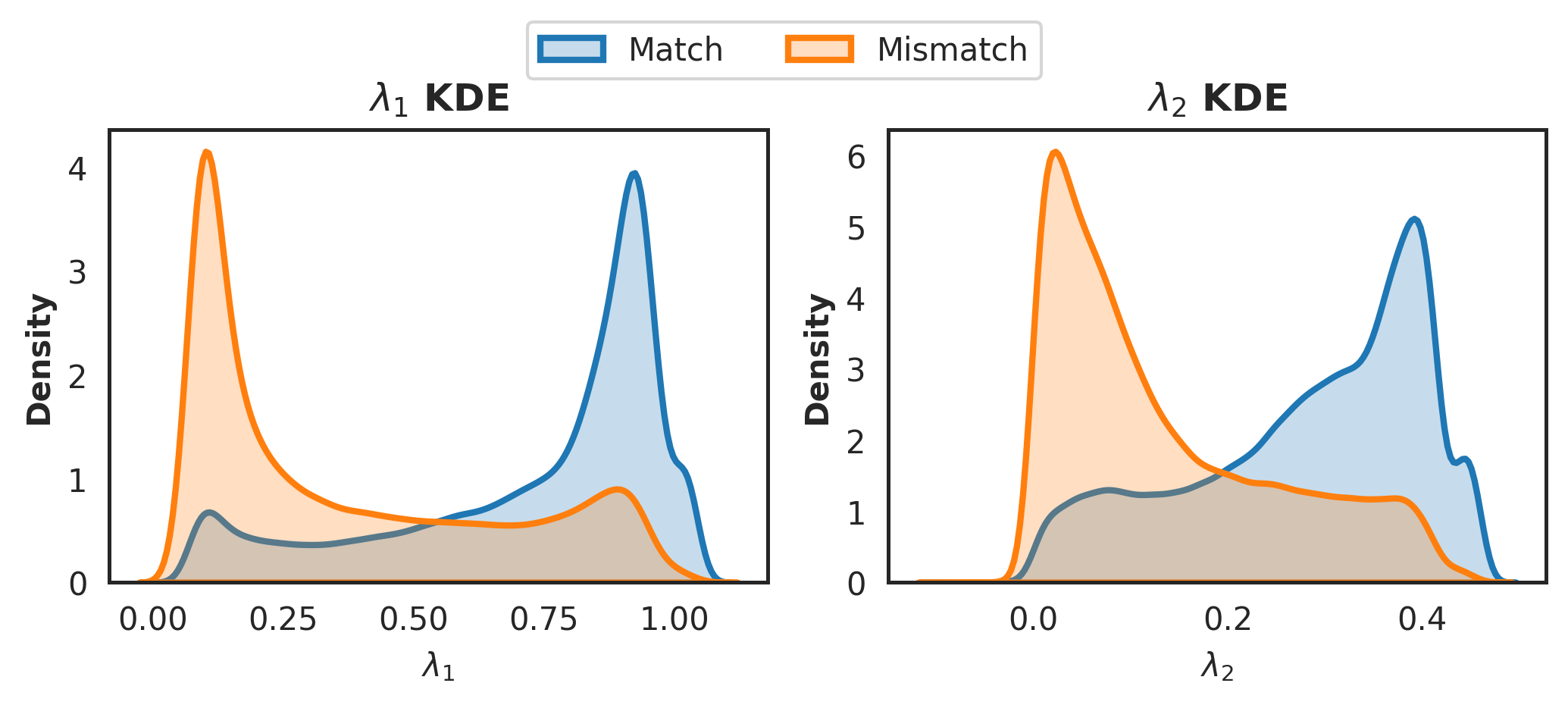}
        \caption{\textsc{Yahoo! Answers}}
    \end{subfigure}

    \begin{subfigure}{0.5\linewidth}
        \centering
        \includegraphics[width=\linewidth]{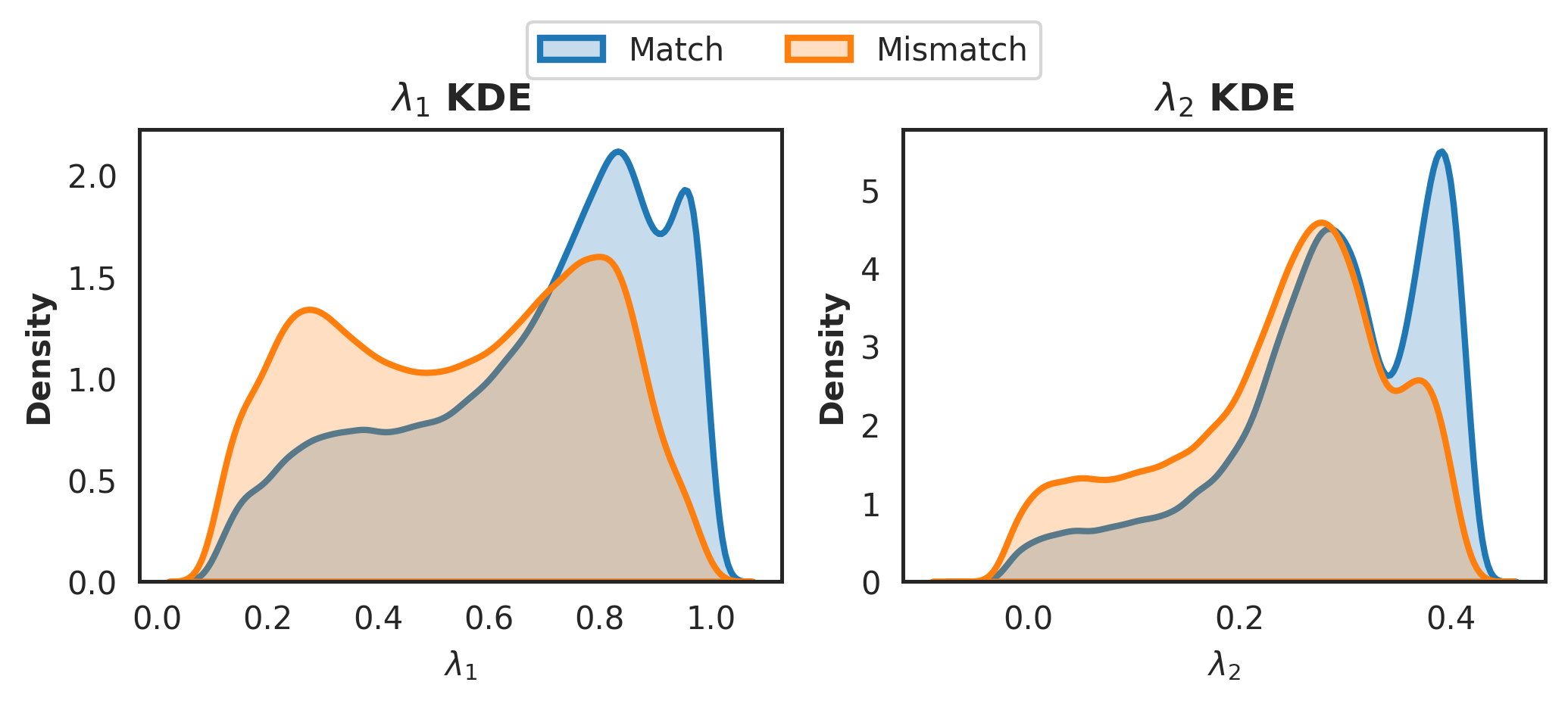}
        \caption{\textsc{Yelp Review}}
    \end{subfigure}

    \caption{Kernel density plots of $\lambda_1$ and $\lambda_2$ weights for label-matched and mismatched samples across five datasets.}
    \label{fig:kde_plots}
\end{figure*}

In this section, we provide kernel density estimates (KDE) of the weighting metrics $\lambda_1$ and $\lambda_2$ across all five benchmark datasets to evaluate the consistency and effectiveness of our sample weighting strategy. For each dataset, we visualize the distribution of weights assigned to samples where the LLM-generated pseudo-labels match the ground truth (Match) and where they do not (Mismatch), as shown in Figure~\ref{fig:kde_plots}.

Across four out of five datasets: AG NEWS, Amazon Reviews, Yahoo! Answers, and Yelp Reviews—we observe a clear separation between the Match and Mismatch distributions. Correctly labeled samples consistently receive higher weights, as shown by the rightward shift of the blue curves, while mislabeled ones are assigned lower weights. These trends align with our overall performance improvements in these datasets.

The only exception is IMDB, where the Match and Mismatch distributions overlap more significantly, suggesting reduced discriminative capacity in weighting. Correspondingly, this is also the only dataset where our method does not outperform LLM baselines, reinforcing the link between effective sample weighting and performance gains.

% \begin{figure*}[t]
%     \centering
%     \includegraphics[width=0.65\linewidth]{plots/aclImdb_kde.png} \\
%     \includegraphics[width=0.65\linewidth]{plots/ag_news_kde.png} \\
%     \includegraphics[width=0.65\linewidth]{plots/amazon_review_kde.png} \\
%     \includegraphics[width=0.65\linewidth]{plots/yahoo_answers_kde.png} \\
%     \includegraphics[width=0.65\linewidth]{plots/yelp_review_kde.png}

%     \caption{Kernel density plots of $\lambda_1$ and $\lambda_2$ weights for label-matched and mismatched samples across five datasets.}
%     \label{fig:kde_plots}
% \end{figure*}

\section{Variability vs. Accuracy}
\label{apx:variability}

\textbf{LG-CoTrain} is designed to exploit complementary behavior between two jointly trained models, which is reflected in the performance gains reported in Table~2 of the main paper. In our framework, each model assigns a per-sample weighting factor $\lambda$ when computing the loss. For model 1, the weight is defined as $\lambda_1 = \text{confidence} + \text{variability}$, while for model 2 it is $\lambda_2 = \text{confidence} - \text{variability}$. This asymmetry in weighting causes the two models to prioritize different examples during training, leading them to learn distinct yet complementary representations. 

To further understand the benefits of this design, we include the \textbf{ST-Random} baseline, where a single model is trained using a randomly selected $\lambda \in \{ \lambda_1, \lambda_2 \}$ for each sample. This baseline consistently underperforms LG-CoTrain, underscoring the importance of the dual-model structure in which each model makes unique and complementary contributions to the final prediction.

\subsection{Distributional Analysis of $\delta_\lambda$}
To quantify the divergence between the two models, we analyze the difference in sample weights:
\[
\delta_\lambda = \lambda_1 - \lambda_2.
\]
A larger range of $\delta_\lambda$ indicates stronger divergence in how the models prioritize samples, suggesting that the two models are being trained differently and thus forming distinct perspectives.

We find that \textbf{Yahoo Answers} and \textbf{Yelp Reviews} exhibit the widest ranges of $\delta_\lambda$ (approximately 0 to 0.8), followed by \textbf{Amazon Reviews} (approximately 0 to 0.6). In contrast, \textbf{AG News} and \textbf{IMDB} show narrower ranges (approximately 0 to 0.2 and 0 to 0.1, respectively), suggesting more similar sample prioritization between the models. 

% Correspondingly, LG-CoTrain achieves the largest performance improvements over baselines on Yahoo, Yelp, and Amazon, while the gains are smaller on AG News and IMDB. These trends support the hypothesis that greater divergence—and thus more unique contributions from each model—leads to improved overall performance.

\begin{figure*}[t]
    \centering

    \begin{subfigure}{0.7\linewidth}
        \centering
        \includegraphics[width=0.6\linewidth]{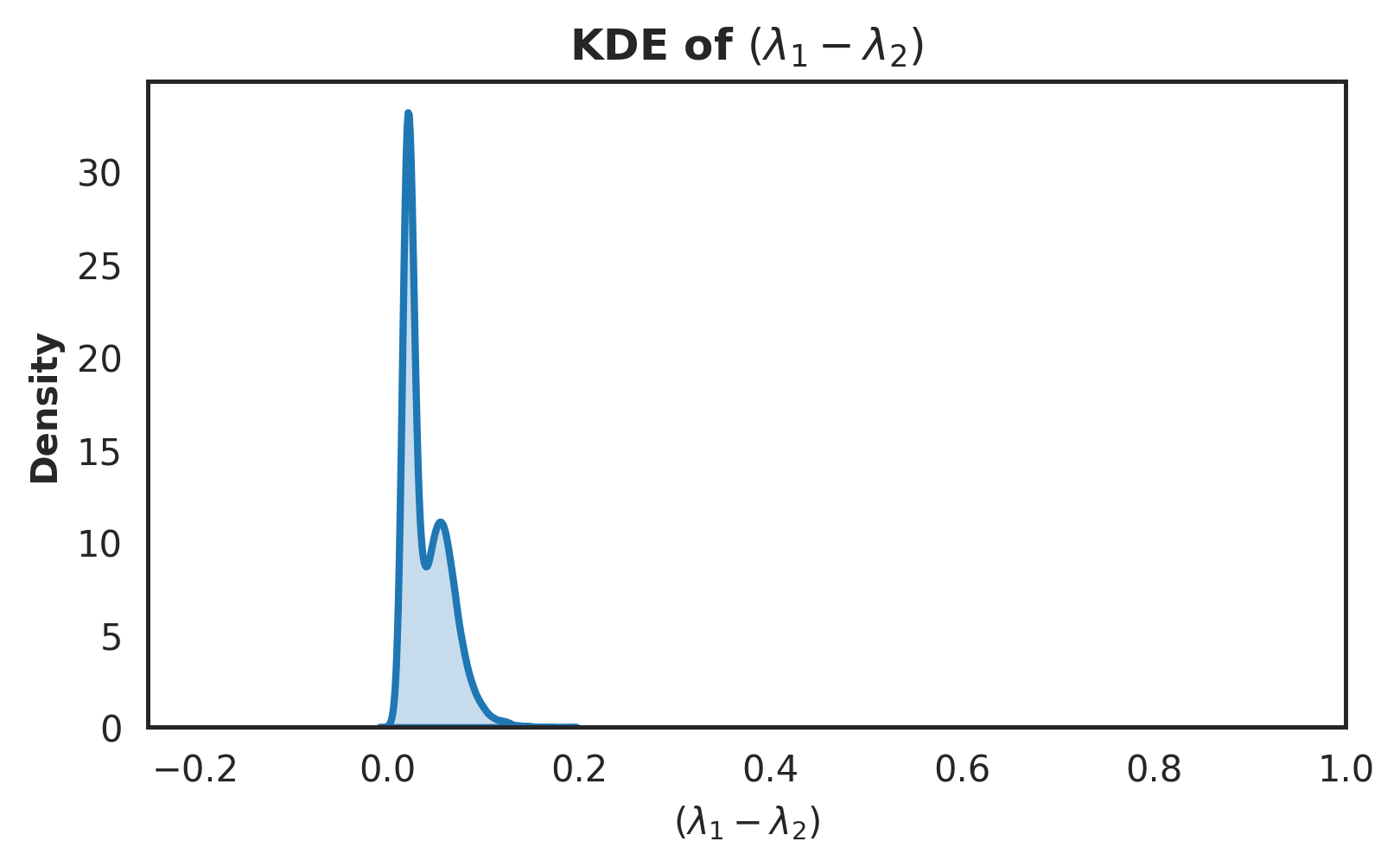}
        \caption{\textsc{IMDB}}
    \end{subfigure}

    \begin{subfigure}{0.7\linewidth}
        \centering
        \includegraphics[width=0.6\linewidth]{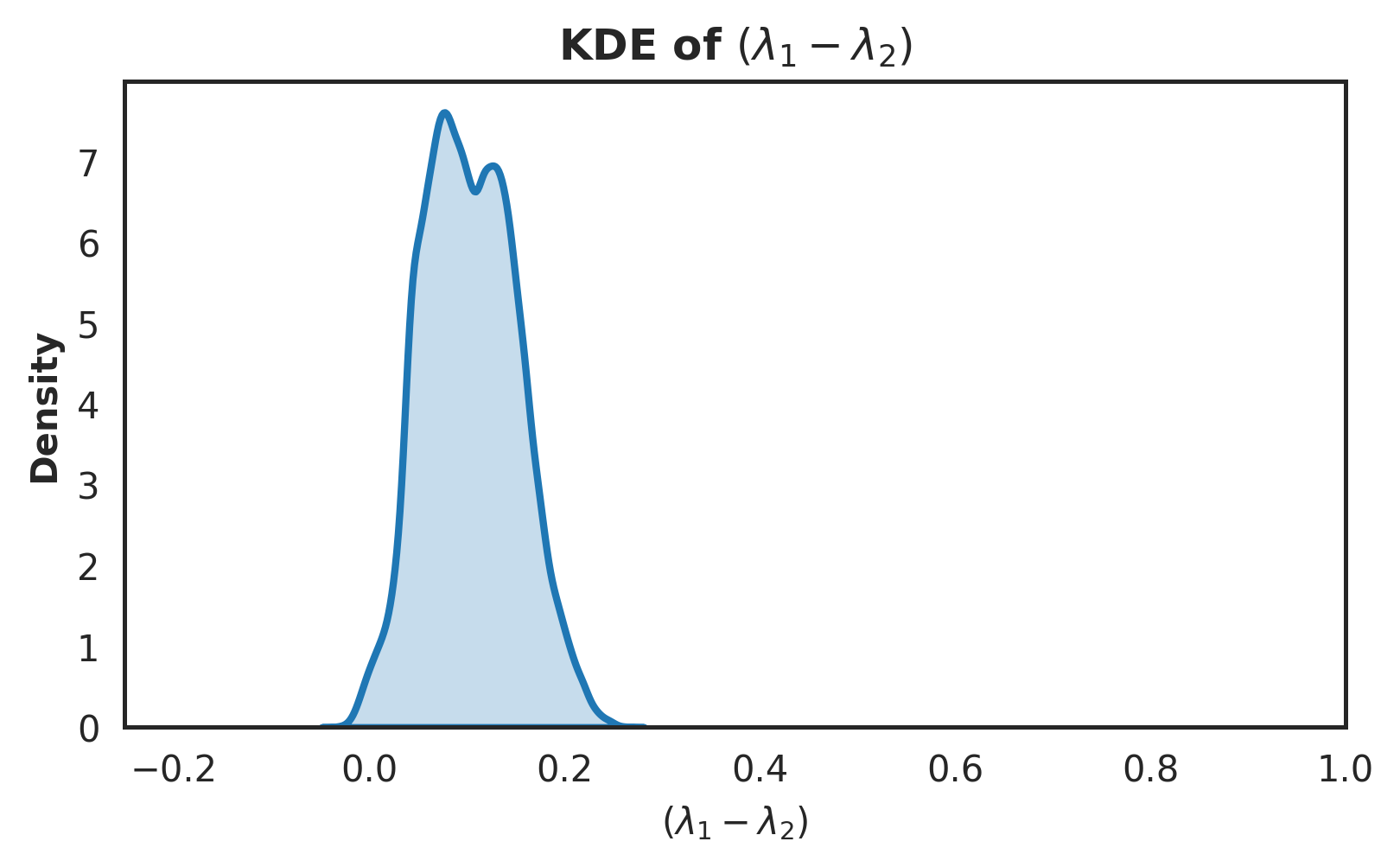}
        \caption{\textsc{AG News}}
    \end{subfigure}

    \begin{subfigure}{0.7\linewidth}
        \centering
        \includegraphics[width=0.6\linewidth]{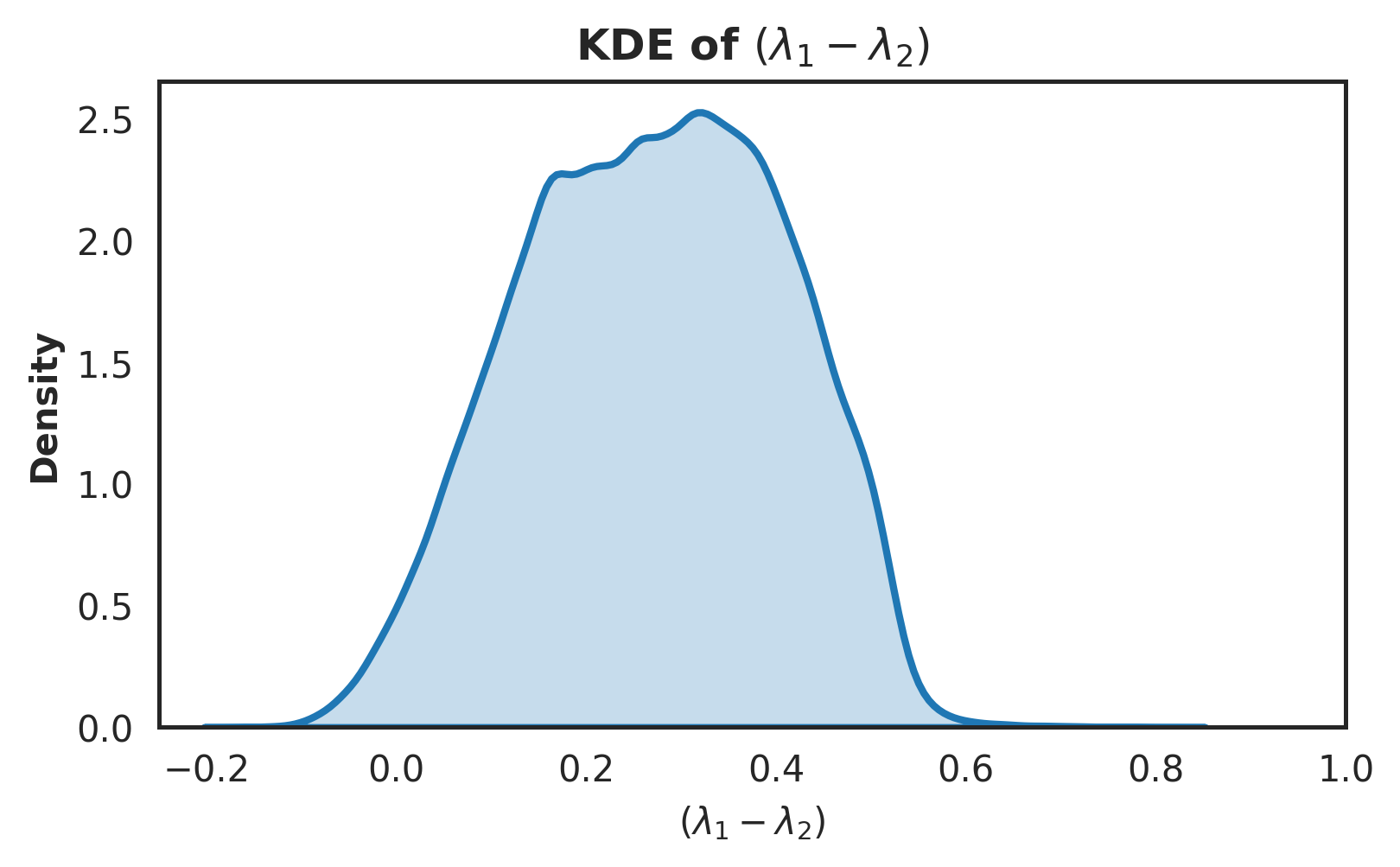}
        \caption{\textsc{Amazon Review}}
    \end{subfigure}

    \begin{subfigure}{0.7\linewidth}
        \centering
        \includegraphics[width=0.6\linewidth]{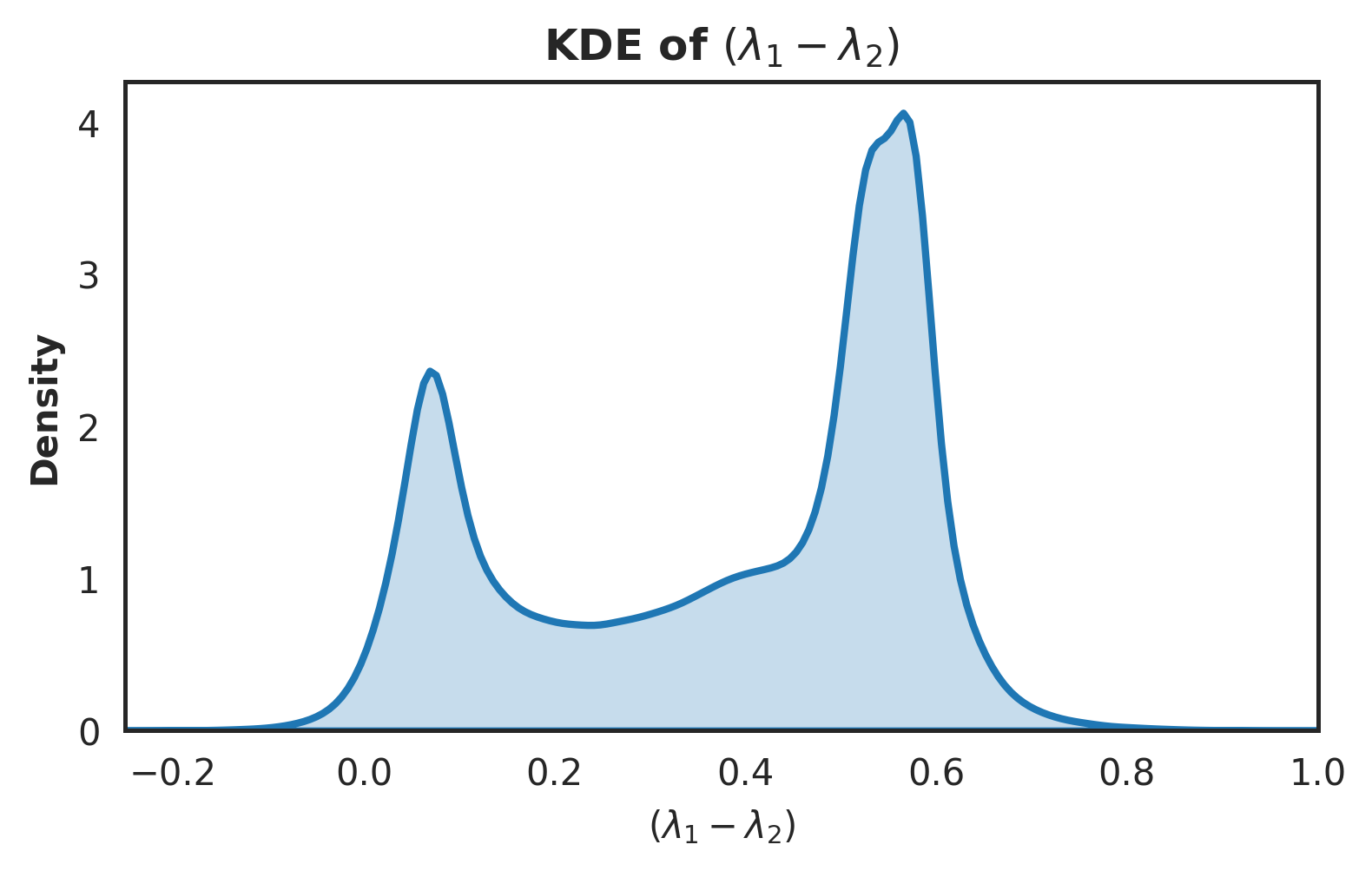}
        \caption{\textsc{Yahoo! Answers}}
    \end{subfigure}

    \begin{subfigure}{0.7\linewidth}
        \centering
        \includegraphics[width=0.6\linewidth]{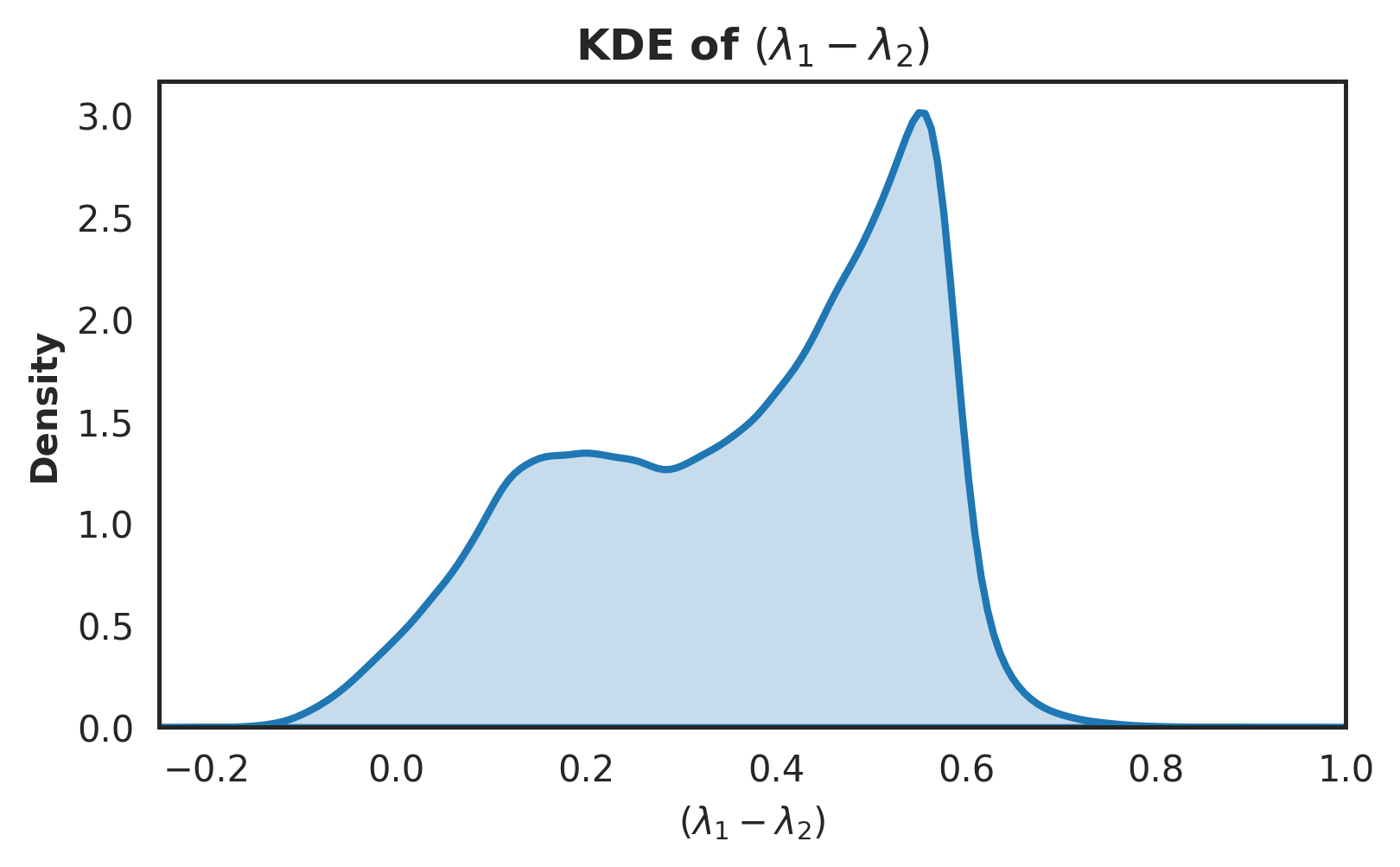}
        \caption{\textsc{Yelp Review}}
    \end{subfigure}

        \caption{Kernel density plots of $\delta_\lambda = (\lambda_1 - \lambda_2)$ for all samples across five datasets.}
    \label{fig:kde_delta_plots}
\end{figure*}

Correspondingly, LG-CoTrain achieves the largest performance improvements over baselines on Yahoo, Yelp, and Amazon, while the gains are smaller on AG News and IMDB. These trends support the hypothesis that greater divergence—and thus more unique contributions from each model—leads to improved overall performance. To better illustrate this relationship, we added Figure~\ref{fig:kde_delta_plots}, which visualizes the distributions of $\delta_\lambda$ across all datasets.

\end{document}